\documentclass[nohyperref]{article}

\usepackage{lipsum}

\usepackage{microtype}
\usepackage{graphicx}
\usepackage{subcaption}
\usepackage{booktabs} 
\usepackage{wrapfig}

\usepackage[ruled, vlined]{algorithm2e} 

\usepackage[backref=page]{hyperref}

\usepackage[accepted]{icml2023}

\usepackage{xargs}

\usepackage{amssymb}
\usepackage{amsmath}
\usepackage{bm}

\DeclareMathOperator*{\argmax}{arg\,max}

\newcommand{\designnet}{\pi_\phi}

\newcommand{\latent}{\theta}

\newcommandx{\histmarg}[1][1=\designnet]{p(h_T| #1)}
\newcommandx{\histlik}[1][1=\designnet]{p(h_T|\latent, #1)}
\newcommandx{\histliki}[2][2=\designnet]{p(h_T|\latent_{#1}, #2)}

\newcommand{\iid}{\overset{\scriptstyle{\text{i.i.d.}}}{\sim}}

\newcommand{\R}{\mathbb{R}}
\newcommand{\E}{\mathbb{E}}

\def\rva{{\mathbf{a}}}

\def\rvc{{\mathbf{c}}}

\def\rvm{{\mathbf{m}}}

\def\rvr{{\mathbf{r}}}

\def\rvy{{\mathbf{y}}}

\def\rvA{{\mathbf{A}}}
\def\rvC{{\mathbf{C}}}

\def\gA{\mathcal{A}}

\def\gL{\mathcal{L}}
\def\gD{\mathcal{D}}

\newcommand{\unconditionaljoint}{p(\rvy, \rvm^* \mid \rvC, \rvC^*, \rvA)}
\newcommand{\unconditionaljointzero}{p(\rvy, \rvm^*_0 \mid \rvC, \rvC^*, \rvA)}

\newcommand{\eig}{\textnormal{EIG}}
\newcommand{\mveig}{\textnormal{CMV-EIG}}

\newcommand{\CApsi}{\psi, \rvC, \rvA}
\newcommand{\CA}{\rvC, \rvA}
\newcommand{\apsi}{\psi, \rva}

\usepackage{amsmath}
\usepackage{amssymb}
\usepackage{mathtools}
\usepackage{amsthm}

\usepackage[capitalize,noabbrev]{cleveref}

\theoremstyle{plain}

\theoremstyle{definition}

\theoremstyle{remark}

\usepackage[textsize=tiny]{todonotes}

\usepackage{tikz}
\usetikzlibrary{fit,arrows.meta,external,math,decorations.pathmorphing}

\icmltitlerunning{CO-BED: Information-Theoretic Contextual Optimization via Bayesian Experimental Design}

\begin{document}

\twocolumn[
%
\icmltitle{CO-BED: Information-Theoretic Contextual Optimization via Bayesian Experimental Design}



\icmlsetsymbol{equal}{*}

\icmlsetsymbol{equal}{*}
\icmlsetsymbol{cmsr}{$\dagger$}

\begin{icmlauthorlist}
\icmlauthor{Desi R. Ivanova}{oxford,cmsr}
\icmlauthor{Joel Jennings}{msr}
\icmlauthor{Tom Rainforth}{oxford}
\icmlauthor{Cheng Zhang}{msr}
\icmlauthor{Adam Foster}{msr}
\end{icmlauthorlist}

\icmlaffiliation{oxford}{Department of Statistics, University of Oxford}
\icmlaffiliation{msr}{Microsoft Research}

\icmlcorrespondingauthor{Desi R Ivanova}{desi.ivanova@stats.ox.ac.uk}
\icmlcorrespondingauthor{Adam Foster}{adam.e.foster@microsoft.com}

\icmlkeywords{Machine Learning, ICML}

\vskip 0.3in
]



\printAffiliationsAndNotice{\icmlcontrib} 

\begin{abstract}
We formalize the problem of contextual optimization through the lens of Bayesian experimental design and propose CO-BED---a general, model-agnostic framework for designing contextual experiments using information-theoretic principles. 
After formulating a suitable information-based objective, we employ black-box variational methods to simultaneously estimate it and optimize the designs in a single stochastic gradient scheme.    
In addition, to accommodate discrete actions within our framework, we propose leveraging continuous relaxation schemes, which can naturally be integrated into our variational objective.
As a result, CO-BED provides a general and automated solution to a wide range of contextual optimization problems. 
We illustrate its effectiveness in a number of experiments, where CO-BED demonstrates competitive performance even when compared to bespoke, model-specific alternatives.
\end{abstract}

\section{Introduction}

Contextual optimization (CO) is an important problem that arises in a wide range of applications, such as drug design \cite{krause2011contextual}, nuclear fusion \cite{char2019offline, chung2020offline}, and robotics \cite{deisenroth2014multi, kupcsik2017model}. 
The goal in such scenarios is to maximize a context-dependent reward function by assigning optimal actions to different contexts.

A concrete example of this problem is a personalized marketing campaign. Here different actions, such as sending marketing materials or discounts for products, are chosen based on context information such as customers' demographics, preferences, and past engagements with the brand.  
The ultimate goal is to maximize revenue, but this first requires us to gather data and learn about customers' behavior. 

We consider a problem setting where we first gather data in an \textit{experimentation stage} that consists of performing actions on a number of different contexts in parallel. At the end of the experiment, data is collected and used to inform a strategy that is then \emph{deployed} (without additional feedback). 
The success of the first stage is judged on the performance of the deployed strategy: better data in the first phase should lead to better decisions and lower regret at deployment time.  


Making best use of resources in the data gathering stage
necessitates experimental design: we want to gather as much useful information as possible for our downstream decision-making in the deployment phase.
Our first contribution is to 
formalize this using an information-theoretic form of Bayesian experimental design \citep[BED, ][]{lindley1956, chaloner1995,mackay1992information}, thereby providing a highly principled framework for choosing designs (or actions) to be optimally informative.

Unfortunately, information-theoretic BED approaches have not previously been applied in the CO setting, or indeed with contextual information more generally.
%
%
Moreover, while substantial recent progress has been made in underlying computational challenges of information-theoretic BED~\citep{foster2019variational,kleinegesse2020minebed,foster2020unified,ivanova2021implicit}, this has generally focused on targeting information gain in model parameters, rather than the contextual optima we are interested in.


Targeting information gain in optima has separately been considered in the Bayesian optimization (BO) literature, where it is commonly referred to as entropy search
\citep[ES,][]{hennig2012entropy,hernandez2014predictive, wang2017max}.
However, these approaches have not been applied in contextual settings, and their usage has been heavily dependent on exploiting model-specific properties to make the required computations tractable; they cannot be directly applied in more general settings.

In this paper, we propose using an information-theoretic BED approach to CO and introduce
CO-BED---a general, model-agnostic framework for designing large-scale contextual experiments.
We begin by formulating a suitable information-theoretic objective, the
contextual max-value EIG (CMV-EIG).
Importantly, CMV-EIG is \emph{transductive}: it measures how much information an observation at one context contains about the optimum at another.
As it represents a mutual information between finite-dimensional random variables,
we can use black-box variational methods to simultaneously estimate CMV-EIG and optimize the designs in a single stochastic gradient scheme.

Gradient-based BED has hitherto been restricted to continuous designs, a significant limitation for contextual optimization where discrete actions are common (e.g.~contextual bandits).
In CO-BED, we therefore propose using the Gumbel-Softmax continuous relaxation \cite{maddison2016concrete,jang2016categorical} to smoothly handle discrete actions.

By framing CO using BED, CO-BED does not sacrifice modelling flexibility to attain computational tractability. It instead offers a general-purpose approach that applies to a wide range of problems in seemingly disparate fields, including contextual bandits \cite{chu2011contextual, agrawal2013thompson, langford2007epoch},  contextual BO \cite{swersky2013multi, ginsbourger2014bayesian, pearce2018continuous, pearce2020practical} and structural equation models \citep{pearl2009causal}.
CO-BED naturally facilitates the design of \emph{large batch} parallel experiments, increasingly a requirement for many applications \citep{groves2018parallelizing, kirsch2019batchbald, zanette2021design, ruan2021linear}.



We demonstrate the benefits of CO-BED in a series of experiments. 
Even when compared against bespoke, model-specific alternatives, we find it consistently performs on par or better, highlighting its effectiveness as a highly applicable and efficient solution.
We further find it is able to scale gracefully, with effective performance maintained on a problem with a $5000$ dimensional design space. 
Our results showcase the promising potential of CO-BED  as an off-the-shelf tool for contextual optimization in various~settings.

\section{Background}
%


\subsection{BED with Expected Information Gain}  \label{sec:background_bed}
Bayesian experimental design \citep[BED,][]{lindley1956} is a principled model-based framework for designing optimal experiments.
BED considers a Bayesian model with experimental outcomes $y$, controllable design $\rva$ and latent parameter $\psi$ with prior $p(\psi)$ and likelihood model $p(y\mid\psi,\rva)$.
The expected information gain (EIG) about the parameters $\psi$ is the expected reduction in entropy from the prior to the posterior distribution of $\psi$ under an experiment with design~$\rva$:
\vspace{-10pt}
\begin{align}
\begin{split}
\eig(\rva) &=  \mathbb{E}_{p(y \mid \rva)} \big[ H[p(\psi)] - H[p(\psi \mid \rva, y)] \big] \\ 
& = \E_{p(\psi)p(y \mid \apsi )} \left[ \log \frac{p(y \mid \apsi)}{p(y \mid \rva)} \right],
\end{split}
\label{eq:EIG}
\end{align}
where $p(y\mid \rva)=\E_{p(\psi)}[p(y \mid \apsi)]$ is the Bayesian marginal distribution of the outcomes and is typically intractable. The EIG is equivalent to $I(\psi; y \mid \rva)$---the mutual information (MI) between the parameters and the experimental outcome under the design.

A common setting, referred to as \emph{batch}, \emph{static} or \emph{fixed} 
\cite{foster2021variational}, is to optimize $D$ designs $\rvA = (\rva_1,\dots,\rva_D)$ simultaneously to maximize the joint information objective, $\eig(\rvA)$. 
By designing and executing informative actions, we collect a dataset $\gD=(\rvA, \rvy)$, which we use to update the model parameters in a Bayesian fashion by computing the posterior $p(\psi \mid \gD)$.
To make inferences about any other quantity of interest, say $\zeta$, we compute its posterior predictive distribution,  $p(\zeta \mid \gD) = \E_{p(\psi \mid \gD)}[p(\zeta \mid \gD, \psi)]$.

\subsection{Black-box MI Estimation and Optimization}\label{sec:background_bbmi}

Despite its highly desirable properties, estimating and maximizing the EIG~\eqref{eq:EIG} is notoriously difficult. This is due to its \emph{doubly intractable} \cite{rainforth2018nesting, zheng2018robust} nature, which is characterized by the nested expectation structure involving a nonlinear function of an intractable term (\citealp[for further details see][]{foster2019variational}). 
To tackle this challenge, we can draw upon recent advances in self-supervised representation learning (\citealp[see][for a review]{poole2018variational}) that have inspired the development of flexible, model-agnostic approaches for the \emph{joint} estimation and optimization of information objectives. 
One such method is based on the InfoNCE lower bound \cite{oord2018representation}, which has been successfully applied in a variety of model-agnostic BED contexts \cite{foster2020unified, ivanova2021implicit, kleinegesse2021gradientbased}, and is given~by
\begin{align}
\begin{split}
    &\eig(\rva) \geq 
    \gL(\rva, U; L) \coloneqq  \\  
    &\E_{p(\psi_{0:L})p(y \mid \rva, \psi_0)} 
    \left[ 
        \log \frac{
            \exp(U(y, \psi_0))
        }
        {
            \frac{1}{L+1}\sum_\ell \exp(U(y, \psi_\ell))
        }
    \right]
\end{split}
\label{eq:infonce}
\end{align}
where $\psi_0 \sim p(\psi)$ is a primary or `positive' sample from the prior, $y \sim p(y \mid \rva, \psi_0)$ is a realisation of the outcome under it, and
$\psi_{1:L} \sim\prod_{i=1}^L p(\psi_i)$ are independent `contrastive' samples. The function  $U: \mathcal{Y} \times \Psi \to \R$ is arbitrary and commonly referred to as a \emph{critic}.
The bound becomes tight in the limit as $L \rightarrow \infty$ for the \emph{optimal} critic $U^*(y, \psi) = p(y \mid \apsi) + c(y)$, where $c(y)$ can be any function that only depends on the outcome $y$. 

If the likelihood $p(y \mid \apsi)$ is analytically available, we can use it in~\eqref{eq:infonce} directly, instead of learning a critic $U$, thus recovering the PCE bound from \citet{foster2020unified}. 
When the likelihood is not analytically available, i.e.~when we are dealing with implicit models, we parameterize $U$ by a neural network with parameters $\phi$ and optimize the lower bound $\gL(\rva, U_\phi; L)$ jointly with respect to $\phi$ and $\rvA$, simultaneously tightening the bound and optimizing the design.

\subsection{Max-value Entropy Search (MES)}\label{sec:background_mes}
The goal of (non-contextual) Bayesian optimization is to find the global maximizer $\rva_\star = \arg\max_{\rva \in\gA} f(\rva)$ of some expensive, black-box function $f$.
Max-value Entropy Search \citep[MES,][]{wang2017max} was proposed as a computationally efficient alternative to earlier methods, such as Entropy Search \citep[ES,][]{hennig2012entropy} and Predictive Entropy Search \citep[PES, ][]{hernandez2014predictive}. 
Whilst ES and PES aim to maximize $I(\rva_\star; y \mid \rva)$---the MI between the outcome under the action queried and the \emph{maximizer}, MES  instead uses the \emph{maximum value}, $m = \max_{\rva \in \gA} f(\rva)$ and maximizes $I(m; y \mid \rva)$ w.r.t.~$\rva$.
The computational efficiency of MES stems from the fact that both $m$ and $y$ are one-dimensional, which reduces the complexity of approximations and makes them more robust and efficient for high-dimensional problems.

Information-theoretic ES methods are popular in large batch BO, as joint MI objectives naturally handle this case. 
MES, like many information-theoretic approaches to BO, focused on
a non-contextual Gaussian process (GP) model \cite{williams2006gaussian} for the black-box function~$f$, allowing for the use of closed-form formulae to approximate the MI. 

\section{Method}

We introduce our approach, CO-BED: Contextual Optimization via Bayesian Experimental Design. At its core, CO-BED seeks to design a set of experiments for exploration purposes, allowing us to gather high-quality data that will lead to better decisions in the subsequent deployment stage.
Code is available at {\scriptsize \url{https://github.com/microsoft/co-bed}}.

\begin{figure}[t]
\centering
    \includegraphics[width=0.43\textwidth]{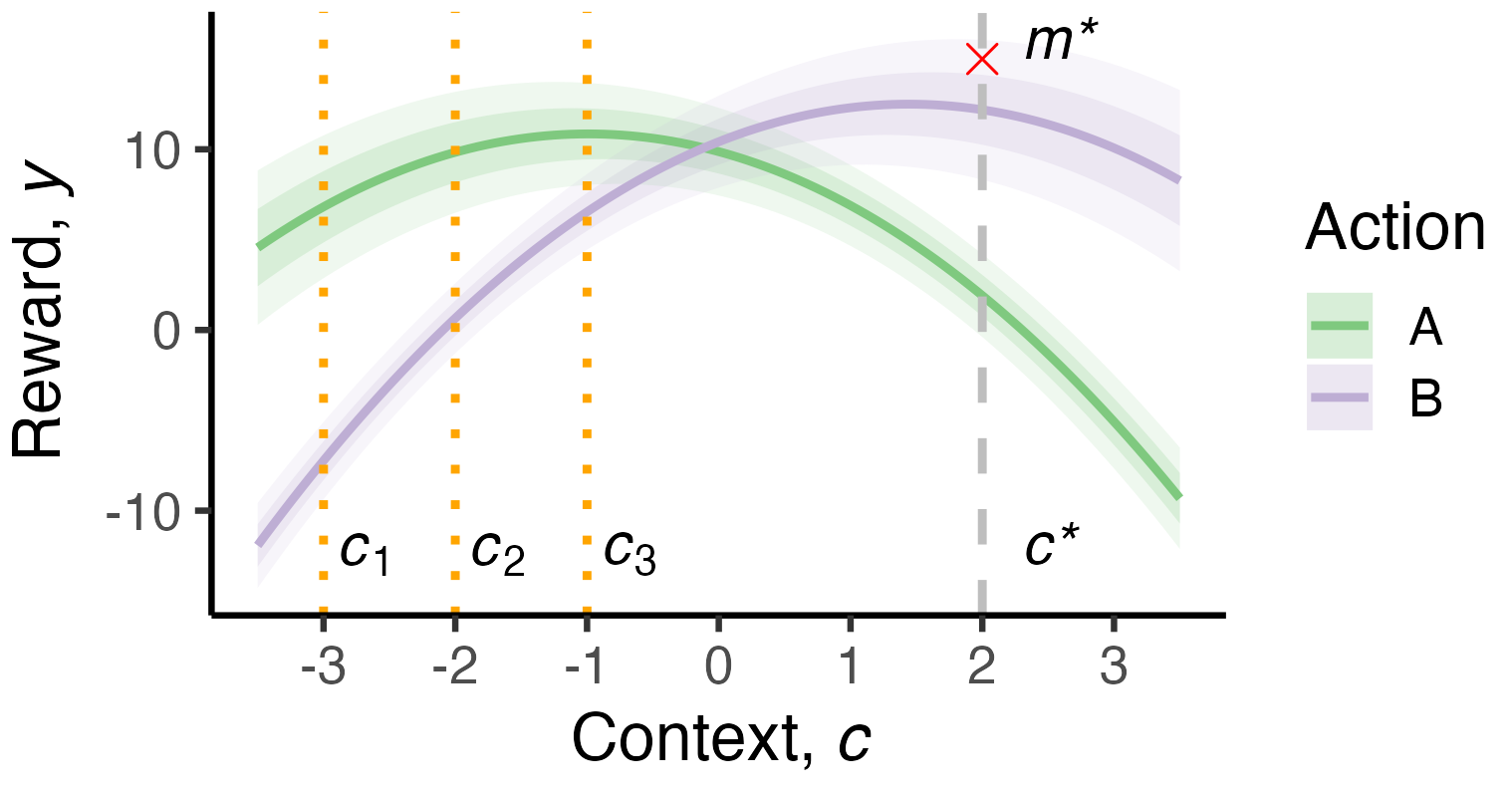}
    \vspace{-10pt}
    \caption{A stylized example. 
    The dark (resp. light) shaded area shows our prior uncertainty about $y \mid \rva,\rvc$ arising from uncertainty in $\psi$, measured by one (resp. two) st.~dev.~from the mean.
    We want to select actions in experimental contexts $\rvc_1, \rvc_2, \rvc_3$ (orange dotted lines) whose outcomes will be most informative about~$\rvm^*$ (red cross) in the evaluation context $c^*$ (grey dashed line).
    \vspace{-15pt}
    }
    \label{fig:stylized_example}
\end{figure}

\textbf{Problem Formulation.}~~
We extend the Bayesian model of \Cref{sec:background_bed} by incorporating a context vector $\rvc$ that is not under the experimenter's control, with the likelihood becoming $p(y \mid \rva,\rvc,\psi)$. Further, we now take $y$ to represent a \emph{reward} with $y \in \R$.
We denote by $m(\rvc) = \E \left[ y \mid \rva_{\star},\rvc \right]$ the maximum value achievable in some context $\rvc$  where $\rva_{\star} = \arg\max_{\rva \in \gA} \E \left[ y \mid \rva, \rvc \right]$ is the action that achieves it. 
Similar in spirit to MES, we wish to learn about these max-values by choosing a large batch of actions to use in an experiment. 
Unlike MES, we want to a) accommodate contextual information these actions are taken under, and b) make our decisions in a \emph{transductive} manner that targets the specific contexts in which our max-values will be evaluated.
Formally, given an externally provided set of \emph{experimental} contexts $\rvC = (\rvc_1,\dots,\rvc_D)$, we seek to design a batch of actions $\rvA = (\rva_1, \dots, \rva_D)$ that
will be maximally informative about the max rewards $\rvm^* = (m(\rvc^*_1),\dots, m(\rvc^*_{D^*}))$ for a given set of \emph{evaluation} contexts $\rvC^* = (\rvc^*_1,\dots,\rvc^*_{D^*})$ which are representative of contexts seen in deployment.

The contexts $\rvC$ and $\rvC^*$ are fixed but arbitrary, so they can be the same, subsets of, or distinct from each other; this is a strict generalization of the standard contextual setting $\rvC=\rvC^*$.
This added flexibility can be essential in practical applications where we  know about the contexts we will encounter at deployment. In the personalized marketing example, the experimental contexts $\rvC$ could represent the customers in a given city who will participate in a real-world experiment, while the evaluation contexts $\rvC^*$ may represent the customers in a whole region where the campaign will be rolled out with the updated model.

We emphasize that the goal of the design process is to obtain data that will aid in learning about the maximum rewards in the evaluation contexts, $\rvm^*$, rather than maximizing the rewards in the experimental contexts. 
This is illustrated in \cref{fig:stylized_example}, where choosing Action~A leads to higher rewards in the experimental contexts, but these rewards will be uninformative about the value of $\rvm^*$, since this action is \emph{a priori} known to be sub-optimal for the context of interest~$\rvc^*$.  
This example also demonstrates that
the typical EIG objective, which aims to reduce uncertainty uniformly across all model parameters $\psi$, is also generally sub-optimal for efficiently learning about max-values of interest. Specifically, spending experimental resources to learn the parameters associated with Action~A would be ineffective and wasteful.   





\begin{algorithm}[t]
\SetAlgoNoLine


\SetKwInput{Input}{Input}
\SetKwInput{Output}{Output}

\textsc{Experimentation phase}

\Input{
Model $p(\psi)p(y, m^*|\psi, \rvc, \rvc^*,  \rva)$, 
initial $\rvA$~and~$U_\phi$,
experimental contexts 
$\rvC$, 
evaluation contexts 
$\rvC^*$
}
\Output{A batch of actions $\rvA^*$ for evaluation contexts $\rvC^*$}

\vspace{2pt}


\textit{Experimental design of $\rvA$}

\Indp

\While{\textnormal{Computational budget not exceeded}}{
$\triangleright$ Sample $\psi \sim p(\psi)$

$\triangleright$ Sample $\rvy, \rvm^*\sim \unconditionaljoint$ 
    
$\triangleright$ Estimate $\gL(\rvA, U_\phi; L)$~\eqref{eq:infonce_mveig} using samples and update the parameters $(\rvA, \phi)$ using SGA
}

\Indm

\textit{Execution of experiment, data gathering and model updating}

\Indp
$\triangleright$ In parallel across $i=1,\dots, D$, for each experimental context $\rvc_i$, apply action $\rva_i$ obtained in previous stage, receive outcome $y_i$. Set $\rvy = \{y_i\}_{i=1}^D$, $\mathcal{D} = \{\rvy, \CA\}$ 

$\triangleright$ Estimate $p(\psi \mid \mathcal{D})$, use it to infer optimal actions $~~\rvA^* = \arg\max_{\rvA^\prime \in \gA^{D^*}} \E_{p(\psi \mid \mathcal{D})} \E \left[ y \mid \rvA^\prime, \rvC^*, \psi \right]$


\Indm

\textsc{Deployment phase}

\Indp
$\triangleright$ Apply $\rvA^*$ to obtain outcomes $\rvy^*$

\caption{CO-BED}
\label{algo:method}
\end{algorithm}



    


\subsection{Contextual Max-value EIG}

Following the principles of information-theoretic BED, we formulate a new objective, the contextual max-value expected information gain (CMV-EIG), for CO.
Our objective focuses on the transductive gain of information about the maximum values $\rvm^*$ in the evaluation contexts $\rvC^*$ when choosing designs $\rvA$ in the experimental contexts $\rvC$:
\begin{align}
\begin{split}
\mveig(\rvA; \rvC, \rvC^*)  &\coloneqq \E
    \left[ 
        \log \frac{  p(\rvy \mid \rvm^*,  \CA) }{ p(\rvy \mid \CA) }
    \right] \\ 
    &=I(\rvm^*;\rvy \mid \rvC, \rvC^*, \rvA).
    \label{eq:mv_eig}
\end{split}
\end{align}
Note that this is equal to the MI between finite-dimensional random variables ($\rvm^*$ and $\rvy$).
The expectation is taken over the joint distribution of outcomes and quantities of interest,  marginalizing out the parameters: $\unconditionaljoint = \E_{p(\psi)} \big[ p(\rvm^* \mid \rvC^*, \psi) p(\rvy \mid \CApsi) \big]$. 
Similarly, the likelihood term is given by $p(\rvy \mid \rvm^*, \CA) = \E_{p(\psi \mid \rvm^*)} [p(\rvy \mid \CApsi)]$, 
and the marginal outcome in the denominator is $p(\rvy \mid \CA) = \E_{p(\psi)} \big[ p(\rvy \mid \CApsi) \big]$. 


By incorporating the concept of experimentation and evaluation contexts, our new objective enables more flexible and targeted experimental design by efficiently allocating resources to learn about the specific contexts of interest. 


\subsection{Lower bounding CMV-EIG}

Our CMV-EIG objective is intractable as none of the likelihood terms involved in~\eqref{eq:mv_eig} are available analytically.
To side-step this, 
we leverage a variational lower bound, which can be optimized with gradients using samples only.
Specifically, utilizing 
an auxiliary \emph{critic} function $U(\rvy,\rvm^*) \in \R$, we adapt the InfoNCE mutual information lower bound introduced by~\citep{oord2018representation} and used in standard implicit-likelihood BED settings by \cite{ivanova2021implicit, kleinegesse2021gradientbased} to our CMV-EIG objective:
\begin{align}
\mathcal{L}(\rvA, U; L) &= 
    \E \left[ 
        \log \frac{ \exp(U(\rvy , \rvm^*_0))}
        {\frac{1}{L+1} \sum_\ell \exp(U(\rvy , \rvm^*_\ell))}
    \right] \label{eq:infonce_mveig} \\
    & \leq \mveig(\rvA; \rvC, \rvC^*),
\end{align}
where the expectation is taken with respect to $\unconditionaljointzero p(\rvm^*_{1:L}\mid\rvC^*)$.
The bound holds for any  number of contrastive samples $L \geq 1$ and critic $U$, and becomes tight as $L \rightarrow \infty$ for the optimal one $U^*(\rvy , \rvm^*) = \log p(\rvy \mid \rvm^*, \CA) + c(\rvy)$, where $c(\rvy)$ is an arbitrary function of~$\rvy$.


The key technical challenge we now face is to find a way to
approximate the expectation~\eqref{eq:infonce_mveig} (or more specifically its gradients) in an unbiased manner. We can do that by generating joint samples $\rvy, \rvm^*\sim \unconditionaljoint=\E_{p(\psi)} \big[ p(\rvm^*\mid \psi, \rvC^*) p(\rvy \mid \CApsi) \big]$,~and contrastive max-values $\rvm^*_{1:L} \sim \prod_{\ell=1}^L p(\rvm^*_\ell \mid \rvC^*)$, where $p(\rvm^*_\ell \mid \rvC^*) = \E_{p(\psi)}[p(\rvm^*_\ell \mid \psi, \rvC^*)]$.
To obtain a sample from the joint, we first sample a parameter $\psi \sim p(\psi)$, then conditionally sample the outcomes $\rvy \sim p(\rvy \mid \CApsi)$ from our model. 

To sample the corresponding max-rewards $\rvm^* \mid \psi \sim p(\rvm^* \mid  \psi, \rvC^*)$, we distinguish three cases. 
First, in certain situations, we might be able to compute $\rvA_\star$ analytically, with many linear and parametric models falling into this category. 
For example, if $\E[y\mid \psi,\rva,\rvc]$ depends linearly on $\rva$, then computing $m^*$ amounts to solving a linear program.
Second, if $\gA$ is discrete, we can determine the optimal actions, $\rvA_\star = \arg\max_{\rvA} \E \left[ y \mid \psi, \rvA, \rvC^* \right]$ by complete enumeration. This captures the majority of the contextual bandit literature.
(Note that the expectation here is only over observation noise, since there is no functional uncertainty.)
Third, when the previous options are not feasible, we choose a finite grid of possible actions, $\tilde{\gA} \subset \gA$, and get an \emph{estimate} of $\rvm^* \mid \psi$ by complete enumeration over $\tilde{\gA}$.

\subsection{InfoNCE lower bound optimization}
Having established how to generate joint samples, we now focus on optimizing $\gL$ with respect to the designs and the critic. 
To do this in practice, we represent the critic as a neural network, $U_\phi$, and optimize its parameters $\phi$ to improve the \emph{tightness} of the bound; optimizing with respect to $\rvA$ improves the \emph{quality} of the designs.
We highlight that, whilst CMV-EIG resembles the MES objective (outlined in~\S~\ref{sec:background_mes}), our approach to determining the optimal designs is quite distinct. 
Concretely, in its estimation procedure, MES first approximates then maximizes the MI directly.
CO-BED never actually computes the MI~\eqref{eq:mv_eig} explicitly, however, provided a sufficiently flexible architecture for $U_\phi$, we expect to obtain tight, high-quality estimates. 

We aim to converge to the true MI maximum by jointly optimizing with respect to $\phi$ and designs $\rvA$ in a single stochastic gradient scheme.
Whilst differentiating with respect to $\phi$ is straightforward, 
taking gradients with respect to $\rvA$ presents a technical challenge due to two reasons: 1) $\rvA$ affects the sampling of the expectation in~\eqref{eq:infonce_mveig}; and 2) unlike network parameters, actions can be continuous or discrete.

\textbf{Continuous action space.}~~ Assuming that the actions are continuous and that the experimental outcomes $y \sim p(\rvy \mid \CApsi)$ are differentiable with respect to $\rvA$, we can form a pathwise gradient estimator \citep{mohamed2020monte} for $\nabla_{\rvA, \phi}\mathcal{L}(\rvA, U_\phi; L)$ 
and optimise it with standard automatic differentiation \citep{baydin2018ad,paszke2019pytorch} and stochastic gradient schemes \citep{kingma2014adam}.

\textbf{Discrete action space.}~~
Previous gradient-based BED methods that utilize variational bounds on MI have primarily focused on fully differentiable models (see~\S~\ref{sec:related_work} for a discussion), and avoided dealing with discrete designs.
Here we propose a simple and practical way to handle discrete actions through the use of Gumbel-Softmax relaxation \citep{maddison2016concrete, jang2016categorical}. This allows us to treat the actions as continuous during the training process and apply pathwise gradients in this case as well.

Suppose we have $K \geq 2$ possible discrete actions, so that we can represent each action $(\rva_d)_{d=1}^D$ as a one-hot vector of size $K$ and $\rvA = (\rva_1, \dots, \rva_D)$ as a $D \times K$ matrix. 
Rather than learning $\rvA$ directly, we introduce a 
distribution over the actions
$\pi_{\bm{\alpha}}$, where $\bm{\alpha}\in \R^{D \times K}$ are trainable parameters, representing the probabilities of selecting each action in each of the $D$ contexts. 
Specifically, during training, the probability of selecting action $k$ in context $d$ is given by:
\begin{align}
    &\pi_{d, k} = \frac{\exp( (\log \alpha_{d, k} + g_{d, k}) / \tau )}{\sum_{j=1}^K \exp( (\log \alpha_{d, j} + g_{d,j}) / \tau )}, 
\end{align}
where $g_{d, k} \sim \text{Gumbel}(0, 1)$ is Gumbel noise and $\tau > 0$ is a temperature hyper-parameter. 
We optimize the parameters $\bm{\alpha}$ and those of the critic network $\phi$ jointly with SGA by sampling $\rvA \sim \pi_\alpha$ and and estimating $\nabla_{\alpha, \phi}\mathcal{L}(\rvA, U_\phi; L)$. At inference time, once the policy is trained, the optimal design for experiment $d$ in the batch is given by $\rva_d = \argmax_k\{\pi_{d, k}\}_{k=1}^K$.


Optimizing $\pi_\alpha$ involves making hyper-parameter choices, notably determining an appropriate value of the temperature $\tau$, and deciding on whether or not to anneal it during training. 
At high temperatures, the estimates of the gradients $\nabla_{\alpha, \phi}\mathcal{L}(\rvA, U_\phi; L)$  tend to be low-variance, providing a strong learning signal for the policy to find good actions $\rvA$. 
Conversely, at low temperatures, gradients tend to exhibit higher variance, but $\pi_\alpha$ is closer to the discrete $\arg\max$ that we use to select the optimal action at inference time. 
We explore these hyper-parameter choices in an ablation study (see~\S~\ref{sec:experiment_ablation}), demonstrating the robustness of our framework to different temperature settings.



\begin{figure*}[t]
  \centering
  \includegraphics[width=0.92\textwidth]{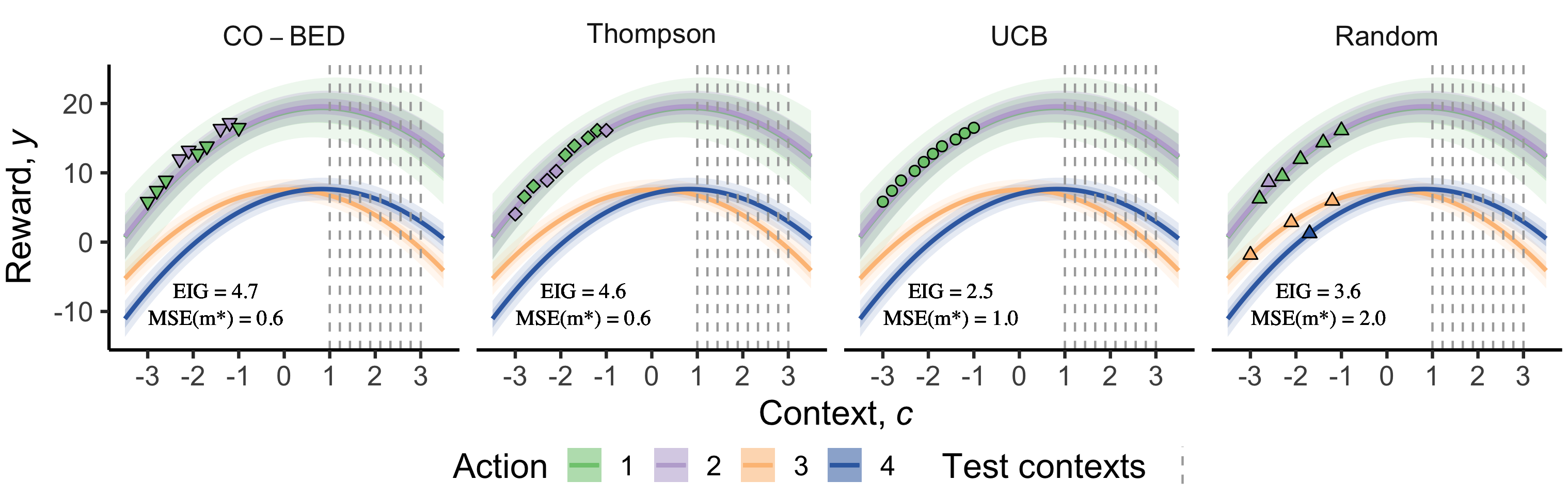}
\vspace{-10pt}
\caption{Discrete actions: designs and key metrics (further available in \S~\ref{sec:appendix_illustrative}). 
The dashed grey lines show the evaluation contexts $\rvC^*$.
}
\label{fig:exp_discrete_ab}
\vspace{-12pt}
\end{figure*}

\section{Related Work}\label{sec:related_work}
CO-BED draws inspiration from several methods across somewhat separate fields to deliver a more general approach to contextual optimization.
Our objective~\eqref{eq:mv_eig} most closely relates to MES for Bayesian optimization \cite{wang2017max}, but differs in two key ways: we focus on CO, instead of finding a single maximizer, and do not use special properties of GPs for MI estimation.
Bayesian Algorithm Execution \citep[BAX, ][]{neiswanger2021bayesian} extends MES beyond global optima to any computable function property.
Rooted in the representation learning literature, InfoNCE~\eqref{eq:infonce_mveig} \cite{oord2018representation,wu2018unsupervised} has been used in BED for non-contextual parameter learning \cite{foster2020unified,ivanova2021implicit,kleinegesse2021gradientbased}.

The problem we address with CO-BED relates to \textbf{contextual Bayesian optimization}, where most of the work to date has focused on iterative acquisition (i.e.~batch size 1) that do not use information-theoretic criteria to choose designs. 
Examples of these methods include Profile Expected Improvement \citep[PEI,][]{ginsbourger2014bayesian}, Multi-task Thompson Sampling \citep[MTS,][]{char2019offline} and knowledge-gradient based methods, such as LEVI, CLEVI, REVI~\citep{pearce2018continuous} and ConBO~\citep{pearce2020practical}.
Many traditional (non-contextual) BO methods have looked at the large batch setting, 
using information-based criteria \cite{hennig2012entropy,wang2018batched}, and alternatives such as local penalization \citep[LP,][]{gonzalez2016batch}, Multi-points Expected Improvement \citep[q-EI,][]{chevalier2013fast}, and the parallel knowledge-gradient \cite{wu2016parallel}.
To the best of our knowledge, the method of \citet{groves2018parallelizing}, combining LEVI and LP, is the only one that considers the large batch, contextual setting and is thus the  only one directly comparable to CO-BED.
\citet{sussex2022model} considered BO in a structural equation model, in a non-contextual case with a known causal graph.


Our method also relates to the broad framework of \textbf{contextual bandits}. 
A significant portion of bandits-related research has focused on the online, linear case \cite{auer2002using, abe2003reinforcement, chu2011contextual, dani2008stochastic, abbasi2011improved, li2019nearly}.
Additionally, some connections with the BO literature have been established with the introduction of Gaussian process bandit optimization methods, such as GP-UCB \cite{srinivas2010gaussian} and its contextual version, CGP-UCB \cite{krause2011contextual}.
More recently, there has been an increased interest in the large batch setting \citep{han2020sequential, ruan2021linear, zhang2021almost}, where the goal is to achieve (some notion of) optimal regret by performing a few rounds of batched experiments.
Closest to our problem set-up is the work of \citet{zanette2021design}, who aim to design a single batch to collect a good dataset that is used to learn a near-optimal policy to be used at deployment time. 
Our approach differs from typical contextual bandits methods in that we focus on information-based criteria, instead of asymptotic regret, and do not restrict ourselves to linear~models.

Our method is the first to consider \emph{contextual} information in the the \textbf{Bayesian experimental design} framework. 
Using variational bounds for EIG estimation in BED for non-contextual \emph{parameter learning} was first proposed in \citet{foster2019variational}.
Approaches that use such bounds and optimize experimental designs using stochastic gradient procedures at the same time have subsequently been developed \cite{foster2020unified,kleinegesse2020minebed,kleinegesse2021gradientbased, foster2021dad, ivanova2021implicit}. 
All of these methods are limited in their ability to deal with \emph{discrete designs} as they either assume fully differentiable models, or resort to gradient-free methods \cite{kleinegesse2020minebed}.

Finally, \textbf{Bayesian active learning} \cite{mackay1992information, houlsby2011bayesian} and Bayesian (active) causal discovery \cite{murphy2001active, tong2001active}, which can be viewed as important special cases of BED, often focus on the large batch setting, which is of particular interest for our method;
notable examples of large-batch methods from the two fields include BatchBALD \cite{kirsch2019batchbald} and CBED \cite{tigas2022interventions}.
Our use of an explicit evaluation context set $\rvC^*$ is akin to \emph{transductive} active learning \citep{mackay1992information,yu2006active,reitmaier2015transductive,wang2021beyond}, in which one seeks data that will improve model predictions at specific inputs.
The focus in active learning is to make accurate predictions, whereas CO-BED addresses the problem of choosing optimal \emph{actions}, accepting prediction uncertainty at certain actions once they are known to be sub-optimal.

%
%
\begin{table*}[t]
\centering
\vspace{-5pt}
\caption{Continuous actions: 40D design, evenly spaced between -3.5 and 3.5, to learn max-values at 39 evaluation contexts equal to the midpoints of the grid. Random$_\sigma$ samples actions from $\mathcal{N}(0, \sigma)$, $\alpha$ in UCB$_\alpha$ is the confidence bound considered. Further details in~\S~\ref{sec:appendix_illustrative}.
}
\label{tab:cts_40d_main}
\vspace{5pt}
\begin{tabular}{lccccc}
    Method             & CMV-EIG $\uparrow$     & MSE$(\rvm^*) \downarrow$       & MSE$(\rvA) \downarrow$       & Regret $\downarrow$    \\
    \toprule
    Random$_{0.2}$     & 5.407 $\pm$ 0.003 & 0.0041 $\pm$ 0.0001 &   0.544 $\pm$ 0.023 & 0.091 $\pm$ 0.002 \\
    Random$_{1.0}$     & 5.798 $\pm$ 0.004 & 0.0024 $\pm$ 0.0002 &   0.272 $\pm$ 0.018 & 0.060 $\pm$ 0.002  \\
    Random$_{2.0}$     & 4.960 $\pm$ 0.004 & 0.0042 $\pm$ 0.0002 &   0.450 $\pm$ 0.021  & 0.090 $\pm$ 0.002  \\
    $\text{UCB}_{0}$   & 5.774 $\pm$ 0.003 & 0.0069 $\pm$ 0.0005 &   0.747 $\pm$ 0.055 & 0.082 $\pm$ 0.002 \\
    $\text{UCB}_{1}$   & 5.876 $\pm$ 0.003 & 0.0030 $\pm$ 0.0002 &   0.338 $\pm$ 0.024 & 0.067 $\pm$ 0.002 \\
    $\text{UCB}_{2}$   & 5.780 $\pm$ 0.004 & 0.0031 $\pm$ 0.0002 &   0.378 $\pm$ 0.031 & 0.069 $\pm$ 0.002 \\
    Thompson           &  6.184 $\pm$ 0.004& 0.0017 $\pm$ 0.0001 &   0.161 $\pm$ 0.007 & 0.051 $\pm$ 0.001 \\
    \midrule
    \textbf{CO-BED}    & \textbf{6.527} $\pm$ \textbf{0.003} &\textbf{ 0.0014} $\pm$ \textbf{0.0001} &  \textbf{0.143} $\pm$ \textbf{0.018} &\textbf{ 0.044} $\pm$\textbf{ 0.001}
\end{tabular}
\vspace{-10pt}
\end{table*}

\section{Empirical Evaluation} \label{sec:experiments}
We compare the performance of \textbf{CO-BED} to a number of baselines across contextual optimization settings including contextual bandits, contextual BO, and causal structure learning. 
We examine continuous and discrete contexts and actions using parametric and GP-based reward~models. 

The baselines that we consider include model-agnostic ones, such as \textbf{random} designs, upper-confidence bound \citep[\textbf{UCB}, ][]{auer2002using} and \textbf{Thompson} sampling \cite{thompson1933likelihood}.
We note that MTS \citep{char2019offline} reduces to pure Thompson sampling in our setting since $\rvC$ is not under the experimenter's control.
We also consider bespoke, model-specific baselines: in the experiment with GPs, we compare against \textbf{LEVI + LP} \citep{groves2018parallelizing}.
For the experiment involving contextual bandits, we compare CO-BED to the Sampler-Planner (\textbf{S-P}) algorithm of \cite{zanette2021design} and all baselines therein.

\textbf{Evaluation metrics.}~~ Our evaluation metrics include the CMV-EIG itself, which we estimate by evaluating~\eqref{eq:infonce_mveig} with the learnt critic and optimized design.
We also consider three evaluation metrics that are useful for assessing the performance of the updated model in the deployment phase: the accuracy of inferring $\rvm^*$ and $\rvA^*$, measured by the MSE between the ground truth and the mean estimate under the posterior $p(\psi \mid \gD)$, and regret from deploying the inferred optimal actions $\rvA^*$ in the evaluation contexts $\rvC^*$.
See \Cref{sec:appendix_regret} for exact details on computing metrics.

\begin{figure*}[t]
  \centering
   \begin{subfigure}{0.49\textwidth}
    \includegraphics[width=\hsize,clip,trim={0 0 0.5cm 0}]{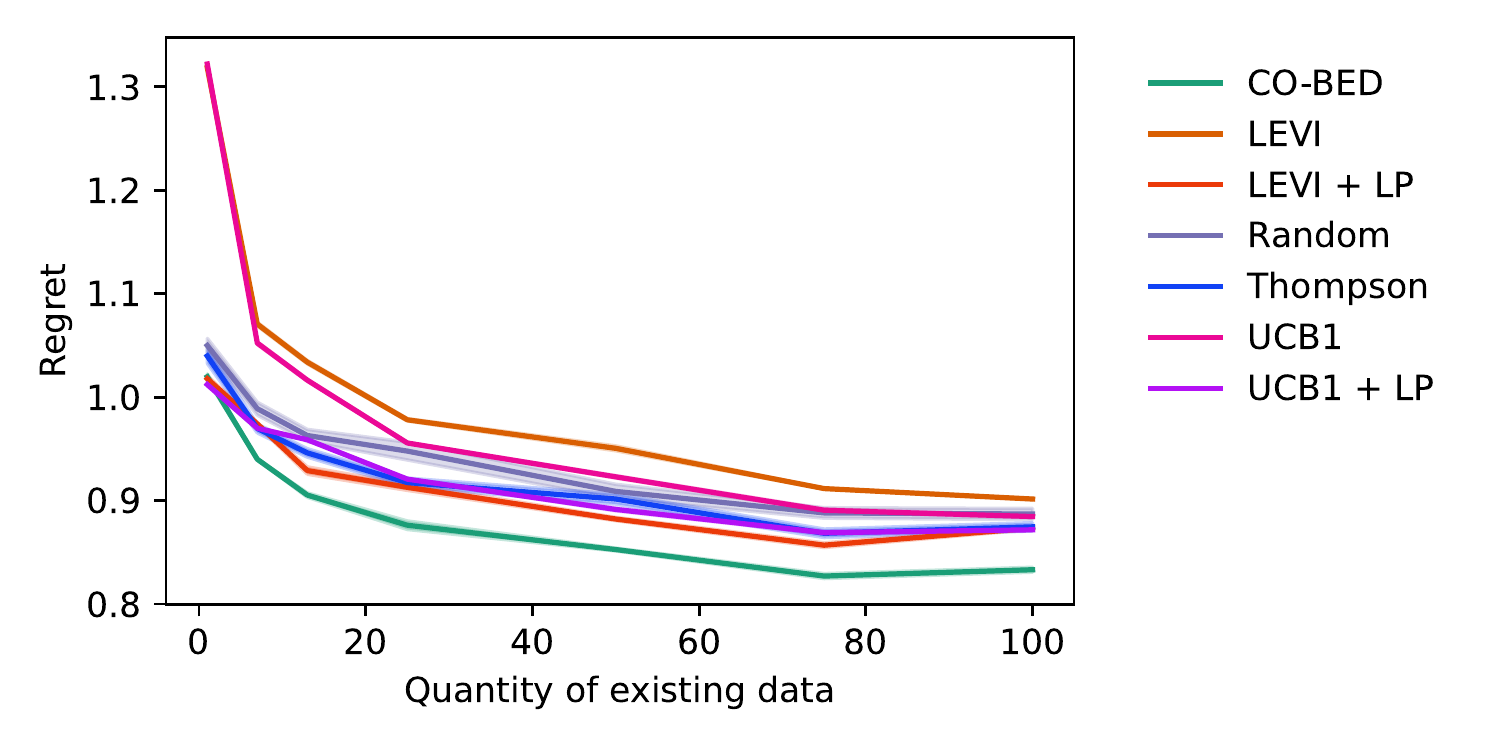}
   \end{subfigure}
    \hfill
   \begin{subfigure}{0.49\textwidth}
       \includegraphics[width=\hsize,clip,trim={0 0 0.5cm 0}]{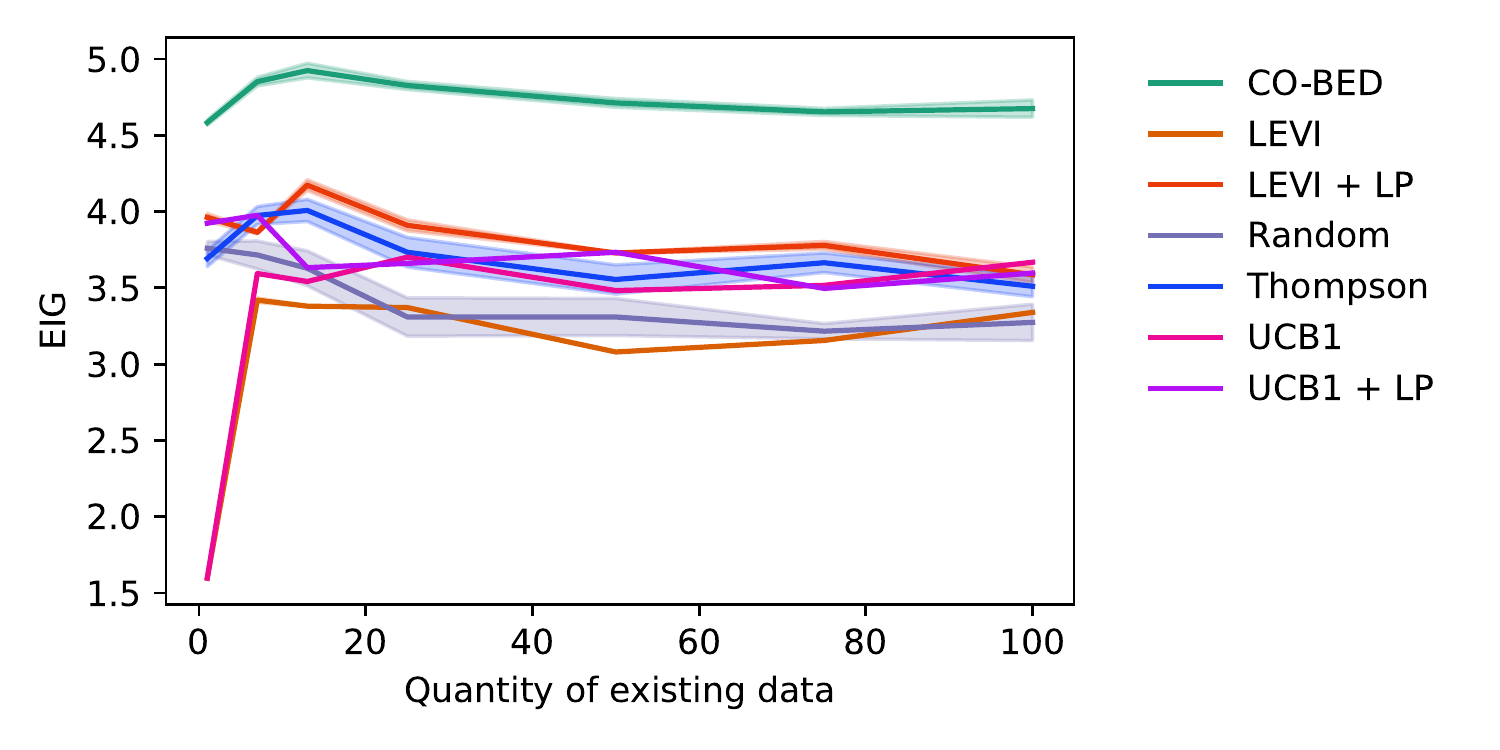}
   \end{subfigure}
   \vspace{-10pt}
   \caption{Results from the Gaussian Process example. Left: the regret evaluated on $\rvC^*$ after experimentation ($\downarrow$ better). Right: the lower bound on the CMV-EIG at the end of training ($\uparrow$ better). Plots show the mean $\pm 1$ s.e.~from 5 seeds. See \Cref{sec:appendix_gp} for details.}
  \label{fig:gp_bias}
\vspace{-10pt}
\end{figure*}

\subsection{Parametric models}\label{sec:experiment_parametric}
We begin our empirical evaluation with two simple parametric models to ensure that our method aligns with intuition and the theory presented in the previous section. Both models have a one-dimensional, continuous context.

The first model we consider has four possible \textbf{discrete actions}, two of which are \emph{a priori} known to be sub-optimal, whilst the other two generate rewards with the same mean, but different variances.
As \cref{fig:exp_discrete_ab} demonstrates, our method has automatically identified the intuitive optimal strategy of A/B testing only the top-performing actions.
Both qualitatively and quantitatively, CO-BED performs on par with Thompson sampling and significantly outperforms the other baselines considered: the random strategy wastes resources by querying sub-optimal treatments, whilst $\text{UCB}_1$ only ever queries the action with higher variance. 

Next, we apply CO-BED to a problem involving \textbf{continuous actions}, designing a batch of 40 experiments to learn about the max-values at 39 evaluation contexts.
The Bayesian model takes the form $y = \exp(-(a-g(\psi, c))^2/h(\psi,c)- \lambda a^2) + \epsilon$, where $g$ and $h$ are parametric functions and $\epsilon$ is Gaussian observation noise, and we can obtain the max values in closed form.
As \cref{tab:cts_40d_main} shows, our method outperforms the baselines on all metrics. 

\subsection{Gaussian Processes}

\label{sec:gp}
We consider modelling the unknown function relating context and treatment to outcomes using a Gaussian Process \citep[GP;][]{williams2006gaussian}. 
We explore the setting in which experimental design begins \emph{after} the acquisition of initial, observational data. We focus on the challenging case of confounded observational data \citep{greenland1999confounding}, in which the context influences the treatment chosen in the observational data. Our experimental designs must learn to counterbalance this bias.
This experiment facilitates a comparison with bespoke GP methods for experimental design. 
We include the method of \citet{groves2018parallelizing} that combines the Local Expected Value of Improvement \citep[LEVI;][]{pearce2018continuous} acquisition function with Local Penalization \citep[LP;][]{gonzalez2016batch}, as one of the few existing approach for batch design for contextual optimization.

Concretely, we let $\rvc \in [-1,1]^2, a \in [-1, 1]$ which are inputs to $\psi: [-1, 1]^3 \to \R$, an unknown function modelled as $\psi \sim \mathcal{GP}(0, k)$. 
We use a simple Gaussian likelihood $y | \rvc, a, \psi \sim N(\psi(\rvc, a), \sigma^2)$ and a radial basis kernel, $k$.
At design time, we condition $\psi$ on a fraction of the 100 observational data points.
At evaluation time, we sample possible ground truth functions $\psi$ from the GP conditional on all 100 observational data (to give consistent evaluation).

Our results in \Cref{fig:gp_bias} show that CO-BED outperforms a wide range of baselines on this problem, particularly with more existing data, demonstrating that CO-BED can learn to deal with a complex prior that is defined by conditioning on confounded observational data.
Baselines using LP do well, but LP's heuristic batch design strategy does not reach the same standard as an information-optimal design.

\subsection{Contextual Linear Bandits}

\begin{figure*}[t]
  \centering
  \includegraphics[width=0.99\textwidth]
  {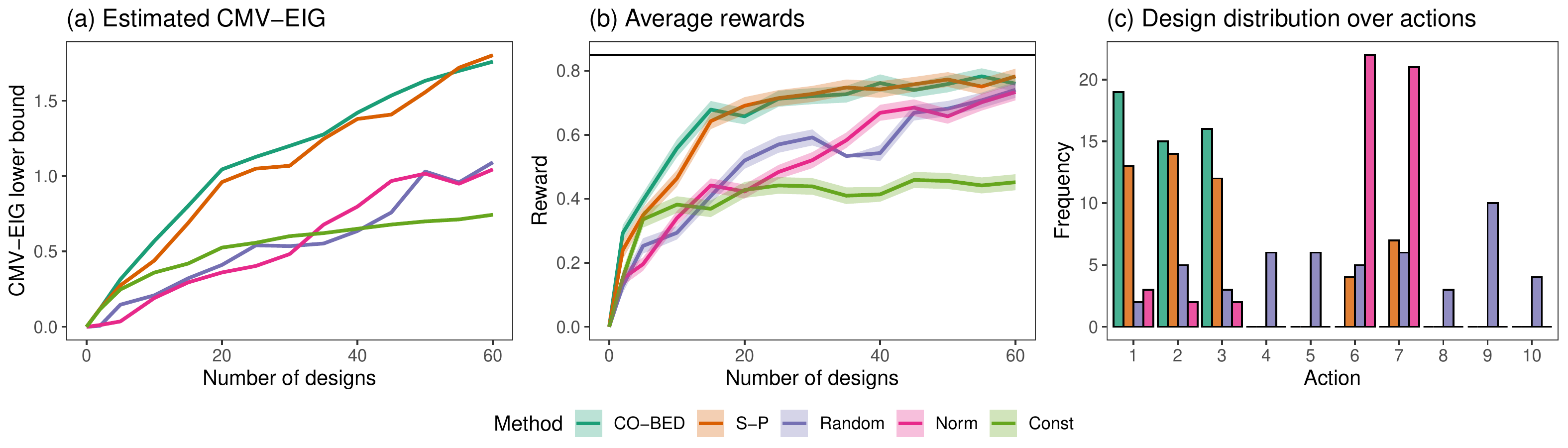}
  ~
  \caption{Linear contextual bandit results. CO-BED performs on par with the bespoke S-P algorithm both from (a) an information standpoint ($\uparrow$ better) and (b) value of the rewards obtained during deployment ($\uparrow$ better). The black horizontal line in (b) represents Supervised Learning (SL), an approximate upper bound on performance. Panel (c) shows the marginal distribution over actions of the designs obtained from each method, excluding Const.}
  \label{fig:linear_bandit}
  \vspace{-10pt}
\end{figure*}

Next, we evaluate our method on the contextual linear bandit problem described in \citet{zanette2021design}, comparing CO-BED to their model-specific Sampler-Planner (S-P) algorithm, as well as the baselines considered therein---a constant strategy (Const), which always chooses action 1, largest norm strategy (Norm), which chooses the feature with the largest norm, and a random design strategy. 
The reward model is defined by
$y= \phi(a, c)^T\psi + \epsilon$, where $\psi = (\psi_1, \dots, \psi_{20})$, is a 20-dimensional parameter vector, $a=1, \dots, 10$ are the possible actions, $c=1, 2, 3$ are the possible contexts, and $\phi: \mathcal{A} \times \mathcal{C} \mapsto \R^{20}$ is a feature map, which is assumed to be known.
The problem is set up so that most actions yield zero average rewards. 
Specifically, only actions 1, 2, and 3 can lead to non-zero rewards in all contexts, which can be used to reduce uncertainty in certain dimensions of the parameter vector, $\psi$. 
Actions 6 and 7 give rise to non-zero features in the last dimension; however, $\psi_{20}$ is essentially zero. All other actions yield exactly zero features.
Full experiment details are given in Appendix~\ref{sec:appendix_exp_bandit}.

\cref{fig:linear_bandit} presents the results, noting that, for consistency with \citet{zanette2021design} we report the average reward, instead of regret.  
CO-BED outperforms the bespoke S-P algorithm at lower ($<15$) design batch sizes and performs on par with it for larger, both in terms of information as well as the value of the rewards obtained during deployment. 
This outperformance is due to its ability to learn the information-optimal designs, as shown in \cref{fig:linear_bandit}(c), specifically in its avoidance of querying actions 6 and 7, thanks to the strong close to~0 prior on $\psi_{20}$. Since S-P does not take prior information into account, it spends some of its experimental resources on those actions, which hurts performance when the experimental budget is low.

\subsection{Unknown Causal Graph}
\label{sec:experiment_causal}

\begin{figure}
    \centering
    \scalebox{0.8}{
	\begin{tikzpicture}
	\node (c1) [draw, circle, fill=lightgray] at (0, 0) {$c_1$};
    \node (cdots) at (0, -0.625) {$\vdots$};
    \node (ckc) [draw, circle, fill=lightgray] at (0, -1.5) {$c_{k}$};
    \node (a1) [draw, circle, fill=lightgray] at (0, -2.5) {$a_1$};
    \node (adots) at (0, -3.125) {$\vdots$};
    \node (aka) [draw, circle, fill=lightgray] at (0, -4) {$a_{\ell}$};
    \node (r1) [draw, circle] at (2, 0) {$r_1$};
    \node (rdots) at (2, -0.625) {$\vdots$};
    \node (rkr) [draw, circle] at (2, -1.5) {$r_{k}$};
    \node (s) [draw, circle, fill=lightgray] at (2, -3.125) {$s$};
    \node (y) [draw, circle, fill=lightgray] at (4, -2) {$y$};
	\draw[-{Latex[length=2.5mm]}] (c1) -> (r1);
	\draw[-{Latex[length=2.5mm]}] (aka) -> (r1);
	\draw[-{Latex[length=2.5mm]}] (ckc) -> (rkr);
	\draw[-{Latex[length=2.5mm]}] (a1) -> (rkr);
    \draw[-{Latex[length=2.5mm]}] (a1) -> (s);
    \draw[-{Latex[length=2.5mm]}] (aka) -> (s);
    \draw[-{Latex[length=2.5mm]}] (r1) -> (y);
    \draw[-{Latex[length=2.5mm]}] (rkr) -> (y);
    \draw[-{Latex[length=2.5mm]}] (s) -> (y);
	\end{tikzpicture}
 }
 \caption{Causal graph considered in \Cref{sec:experiment_causal}. The existence of edges from $a_j$ to $r_i$ is unknown. Unfilled nodes are not observed.}
 \label{fig:causal_graph_tikz}
\vspace{-10pt}
\end{figure}
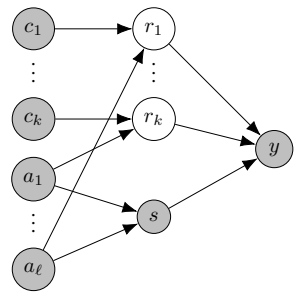

Finally, we explore our method in the context of a structural equation model \citep{pearl2000models,pearl2009causal}.
We look at contextual optimization in a business-inspired scenario with an \emph{unknown} causal graph.
We specifically consider a binary context vector $\rvc \in \{0, 1\}^{k}$ which indicates which business areas a customer is active in, and treatments $\rva \in [0, 1]^{\ell}$ representing investment in different promotional activities.
The unobserved variable $\rvr \in \mathbb{R}^k$ indicates the revenue generated in each business area. 
We related these quantities using a structural equation model with a partially unknown causal graph.
The unknown component of the graph describes which treatments effect which revenue streams, concretely we assume the structural equations $r_i = c_i\sum_{j=1}^{\ell} G_{ij} \theta_{ij}  a_j$ where $G_{ij}$ is a binary matrix and $\theta_{ij}$ are unknown linear coefficients. The total cost of treatments is simply $s = \sum_{j=1}^{\ell} a_j$, and the total observed profit is $y = \sum_{i=1}^k r_i - s + \epsilon$ where $\epsilon$ is Gaussian noise.
The whole system is summarized in \Cref{fig:causal_graph_tikz}.

\begin{figure}
\includegraphics[width=\columnwidth,clip]{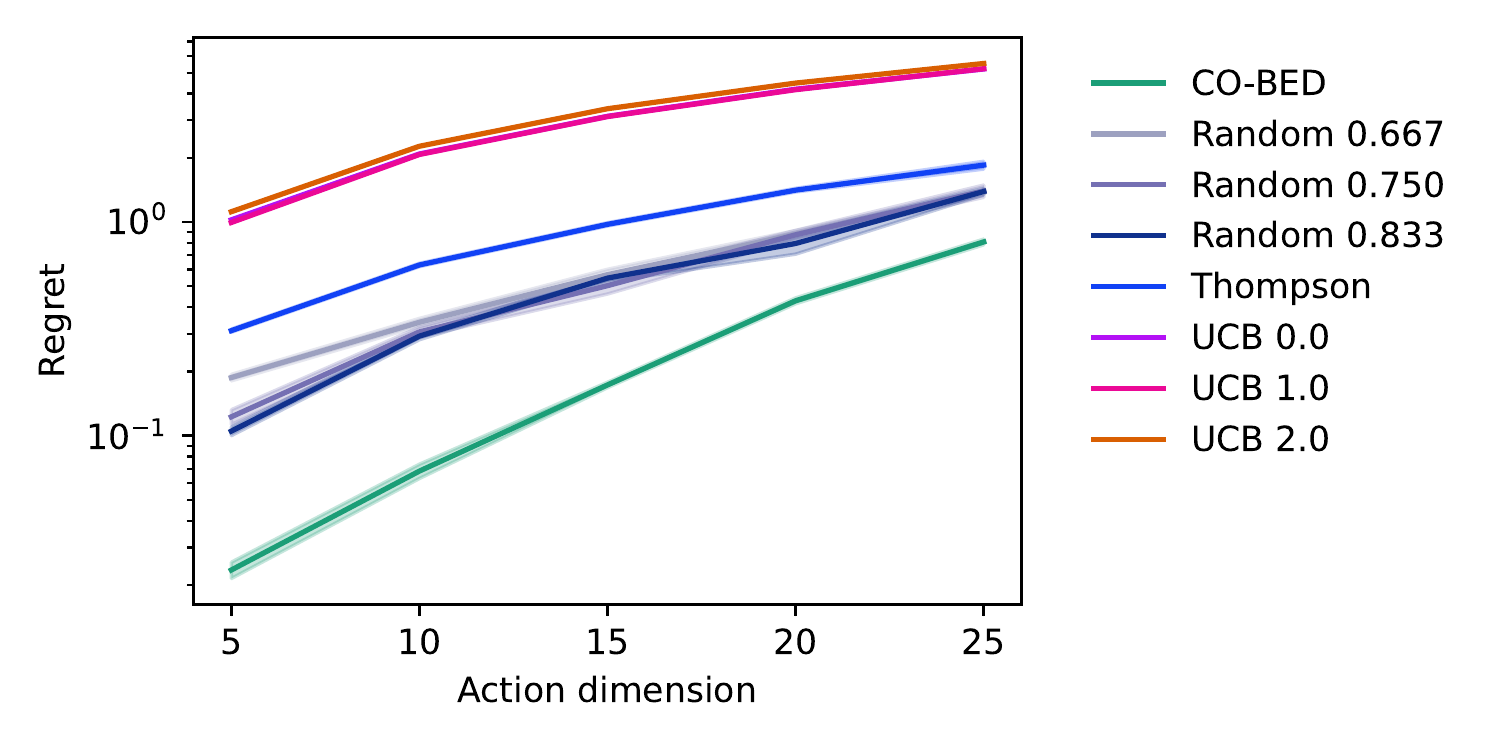}
\vspace{-22pt}
\caption{Results from the Unknown Causal Graph experiment showing regret on $\rvC^*$ after experimentation designed with different methods ($\downarrow$ better). Plots show mean $\pm 1$ s.e.~over 5 seeds.}
\label{fig:causal_graph_regret}
\vspace{-10pt}
\end{figure}

This example also allows us to explore the scalability of our method. We use a fixed number of experimental contexts $D=200$ and vary the number of possible actions $\ell$ up to $25$, yielding designs of up to 5000 dimensions to optimize. 
For evaluation, we let $\rvC^*$ consist of all $2^k - 1$ non-zero binary contexts, and estimate the optimal treatments given observation $\rvy$ by fitting a Lasso \citep{tibshirani1996regression} due to the infeasability of Bayesian inference in this case. See \Cref{sec:appendix_causal} for complete details.
Our results in \Cref{fig:causal_graph_regret} show that our method is successful on this larger problem and outperforms a range of baselines. In particular, UCB baselines struggle here as they do not introduce heterogeneity between treatments.

\begin{figure*}[t]
\includegraphics[width=0.96\textwidth]{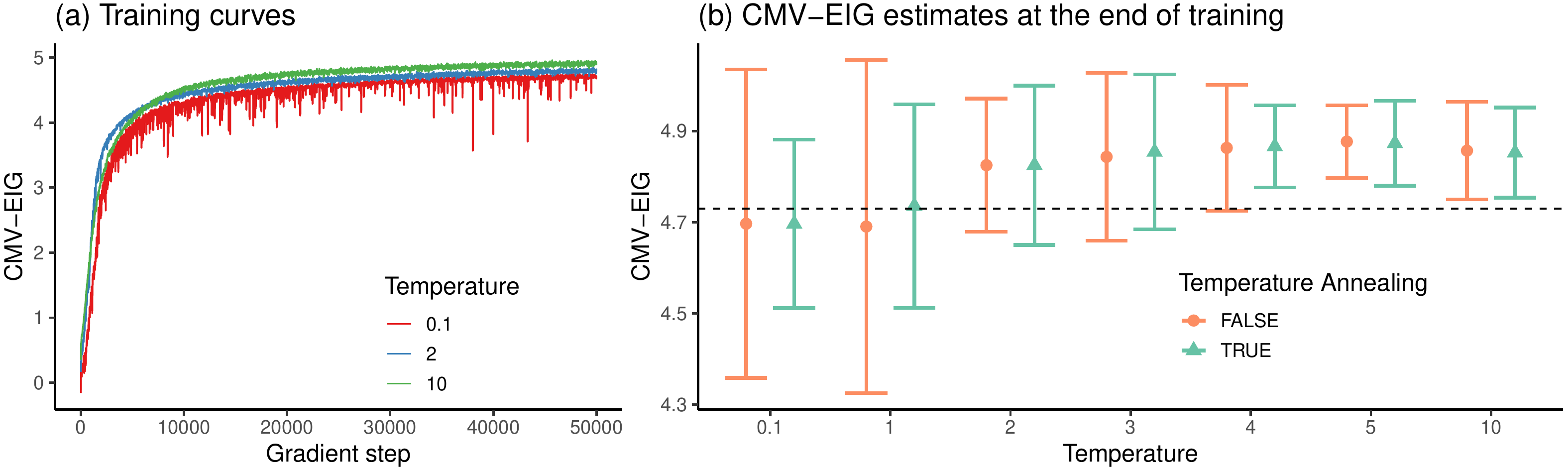}
\vspace{-10pt}
\caption{
Robustness of the Gumbel-Softmax relaxation scheme. Panel (a) shows moving average estimates of CMV-EIG~\eqref{eq:infonce_mveig} (window size of 25) against gradient steps for different \textit{fixed} values of $\tau$ (i.e.~no annealing).
Panel (b) presents CMV-EIG estimates computed at the end of training across 32 initializations of the training parameters (i.e. different seeds). The mean values are marked by green triangles and red dots for scenarios with and without temperature annealing, respectively. Annealing is applied every 10K steps with a factor of $0.5$. Error bars indicate $\pm 2$~st.dev., computed over the 32 seeds. The black dashed line indicates the CMV-EIG estimate reported in \S~\ref{sec:experiment_parametric} achieved by CO-BED, which was trained with temperature annealing and initial $\tau=2.0$.
}
\label{fig:training_curves}
\vspace{-10pt}
\end{figure*}

\subsection{Robustness of the Gumbel-Softmax relaxation} \label{sec:experiment_ablation}

We perform a series of ablation studies to investigate the overall robustness of the Gumbel-Softmax relaxation scheme. We focus particularly on how different choices surrounding the temperature parameter $\tau$ can affect the performance of our framework.
All experiments are performed using the simple parametric model from  \S~\ref{sec:experiment_parametric}.

Figure~\ref{fig:training_curves}(a) shows the training curves of the CMV-EIG lower bound~\eqref{eq:infonce_mveig} for three different temperature values. 
As anticipated, there are noticeable training instabilities at low temperature values ($\tau=0.1$), which stem from the high variance in the estimates of the gradient $\nabla_{\alpha, \phi}\mathcal{L}(\rvA, U_\phi; L)$.
In contrast, both moderate and high temperatures ($\tau=2.0$ and $\tau=10.0$) yield stable training trajectories. 
Importantly, despite these variations, all three settings ultimately converge towards similar values.

Next, we assess the stability across multiple training runs by optimizing $\mathcal{L}(\rvA, U_\phi; L)$ over 32 different seeds, showing the results in Figure~\ref{fig:training_curves}(b).
As expected, variability is larger at lower temperature values, which also tend to result in lower CMV-EIG estimates. 
Nevertheless, all of the CMV-EIG mean values fall within a relatively tight range between 4.7 and 4.9.
However, it is worth noting that the larger CMV-EIG estimates at higher temperatures might be a byproduct of using \textit{soft designs} during training, where we sample $\rvA \sim \pi_\alpha$, but use the discrete $\arg\max$ during deployment.
Thus, soft training  with high temperature values can potentially introduce a subtle train-test mismatch and overestimate CMV-EIG. 
This issue can be resolved by employing hard training or by relying on additional evaluation metrics such as regret and various accuracy metrics (as we do in the experiments section).
Finally, Figure~\ref{fig:training_curves}(b) also suggests that temperature annealing does not significantly affect performance but can help improve training stability, particularly at low temperatures. 


\section{Discussion}
\textbf{Limitations.}~
CO-BED offers a high degree of generality as it applies to a wide range of contextual optimization problems. 
However, this generality comes at the cost of increased computational cost, as it involves learning optimal actions by maximizing a lower bound on the CMV-EIG objective that requires training a (small) neural network.
In many real-world applications, however, this cost is small relative to the overall cost of the experiment which may take several months to run, e.g. in marketing or medical scenario.
Additionally, although common in the BO and bandits literature, future work could investigate ways to lift the assumption that $y$ is continuous and explore more efficient ways to cheaply compute or approximate the conditional max-values $\rvm^* \mid \rvC^*, \psi$.
Finally, in our experiments, we considered a scenario close to \citet{zanette2021design} involving one round of experimentation followed by one round of deployment, but there is no conceptual reason to prevent multiple,~adaptive~rounds.





\textbf{Conclusions.}~ 
We introduced CO-BED---the first method to introduce contextual aspects in the field of BED and to formally connect it to contextual optimization.
By taking an information-theoretic approach, CO-BED offers a general-purpose framework that unifies seemingly disparate fields into a single cohesive framework. 
Our method can be end-to-end trained with gradients by employing black-box variational methods to simultaneously estimate our proposed CMV-EIG objective and optimize the designs in a single stochastic gradient scheme. 
Given the importance of discrete actions in optimization settings, we introduce an approach using the Gumbel-Softmax trick to handle them smoothly. 
We demonstrated the flexibility and effectiveness of our method across a variety of experiments, performing on par or outperforming alternative, bespoke strategies.


\vspace{-2pt}
\section*{Acknowledgements}
\vspace{-3pt}
DRI is supported by EPSRC through the Modern Statistics and Statistical Machine Learning (StatML) CDT programme, grant no. EP/S023151/1.

\newpage
\newpage
\bibliography{refs}
\bibliographystyle{icml2023}

\newpage
\appendix
\onecolumn

\section{Experiments} \label{sec:appendix_experiments}

We implement all experiments in Pyro \citep{pyro}, which is a probabilistic programming framework on top of PyTorch \cite{paszke2019pytorch}.
Our code will be open-sourced upon publication.

\subsection{Computing evaluation metrics}
\label{sec:appendix_regret}
Once we have an experimental design $\rvA$, we simulate the deployment phase of our main set-up (\Cref{algo:method}) to evaluate how well experimental data using $\rvA$ enables us to perform at test time.

We begin by sampling a ground-truth parameter $\psi_\text{true}$. 
We then sample experimental data $\rvy \mid \psi_\text{true}, \rvC,\rvA$.
The knowledge of the experimenter at this point is encapsulated in the posterior $p(\psi \mid \rvC,\rvA,\rvy)$.
Under this posterior, we can then estimate optimal actions and optimal achievable outcomes for the evaluation contexts $\rvc^*_1, \dots,\rvc^*_{D^*}$ via
\begin{align}
    \rva_{\text{post}, i}^* &= \argmax_{\rva' \in \mathcal{A}} \E_{\psi \sim p(\psi \mid, \rvC,\rvA,\rvy)}\left[\E[y \mid \rva', \rvc_i^*,\psi]\right] \\
    m_{\text{post}, i}^* &= \max_{\rva' \in \mathcal{A}} \E_{\psi \sim p(\psi \mid, \rvC,\rvA,\rvy)}\left[\E[y \mid \rva', \rvc_i^*,\psi]\right].
\end{align}
These can be compared with their counterparts under $\psi_\text{true}$
\begin{align}
    \rva_{\text{true}, i}^* &= \argmax_{\rva' \in \mathcal{A}} \E[y \mid \rva', \rvc_i^*,\psi_\text{true}] \\
    m_{\text{true}, i}^* &= \max_{\rva' \in \mathcal{A}} \E[y \mid \rva', \rvc_i^*,\psi_\text{true}];
\end{align}
giving us the `treatment recovery' and `reward recovery' MSEs, which are
\begin{align}
    L_\text{treatment} = \frac{1}{D^*} \sum_{i=1}^{D^*}\|\rva_{\text{post}, i}^* - \rva_{\text{true}, i}^*\|^2 \\
    L_\text{reward} = \frac{1}{D^*} \sum_{i=1}^{D^*}|m_{\text{post}, i}^* - m_{\text{true}, i}^*|^2.
\end{align}
Finally, we evaluate the regret under $\psi_\text{true}$ from acting with $\rvA^*_\text{post}$ as opposed to $\rvA^*_\text{true}$. This is defined as
\begin{equation}
    r_i = m^*_{\text{true},i} - \E[y \mid \rva^*_{\text{post}, i}, \rvc^*_i, \psi_\text{true}] \qquad
    r = \frac{1}{D^*} \sum_i r_i.
\end{equation}
To give a less biased evaluation, this procedure is repeated for several thousand ground truth parameters $\psi_\text{true}$ and the results are averaged. The exact number of true models considered is given in the following sections.

\subsection{Parametric models} \label{sec:appendix_illustrative}

\paragraph{Training details}
All experiment baselines ran for 50K gradient steps, using a batch size of 2048. We used the Adam optimiser \citep{kingma2014adam} with an initial learning rate of 0.001 and exponential learning rate annealing with a coefficient of 0.96 applied every 1000 steps. We used a separable critic architecture \citep{poole2018variational} with simple MLP encoders with ReLU activations and 32 output units. 

For the discrete treatment example:  we added \emph{batch norm} to the critic architecture, which helped to stabilise the optimisation. We had one hidden layer of size 512.
Additionally, for the Gumbel--Softmax policy, we started with a temperature $\tau=2.0$ and \texttt{hard=False} constraint.
We applied temperature annealing every 10K steps with a factor of 0.5. We switch to \texttt{hard=True} in the last 10K steps of training.

For the continuous treatment example: We used MLPs with hidden layers of sizes $[\text{design dimension}\times 2; 412; 256]$ and 32 output units. 

Note: In order to evaluate the EIG of various baselines, we train a critic network for each one of them with the same hyperparameters as above.

\paragraph{Posterior inference details}

\begin{wrapfigure}[15]{r}{0.30\textwidth}
\vspace{-10pt}
  \centering
  \includegraphics[width=0.30\textwidth]{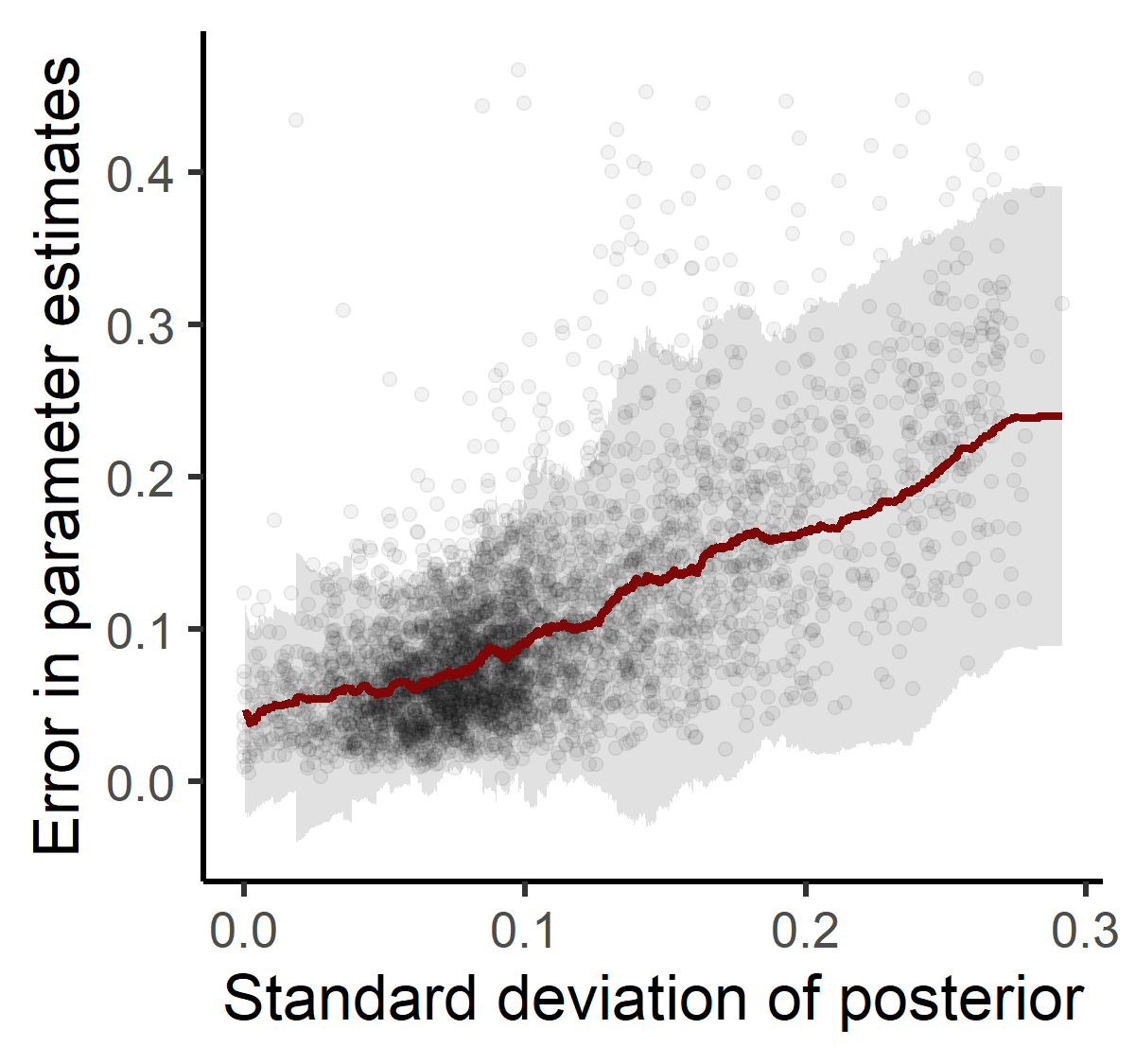}
    \caption{Posterior checks}
    \label{fig:calib}
\end{wrapfigure}

After completing the training stage of our method (Algorithm~\ref{algo:method}), we need to deploy the learnt optimal designs in the real world in order to obtain rewards $\rvy$. This experimental data is then used to fit a posterior $p(\psi|\mathcal{D})$. 

There are many ways to do the posterior inference and the quality of the results will crucially depend on the accuracy of the fitted posteriors. In both of our examples and for all baselines we use Pyro's self-normalised importance sampling (SNIS). Samples from this posterior are used for the evaluation metrics.

We validate the accuracy of estimated posteriors by running various sanity checks, including diagnostic plots such as Figure~\ref{fig:calib}, showing the standard deviation of our posterior mean estimate (a measure of uncertainty about the parameter) and $L_2$ error to the true parameter. The red line shows the rolling mean over 200 points of the latter, and the grey band---the 2 standard deviations. For this plot we used the example of the continuous action with $D=20$ experimental contexts.

\paragraph{Evaluation metrics details}
As discussed in the main text, we evaluate how well we can estimate $\rvm^*$ by sampling a ground truth $\tilde{\psi}$ from the prior and obtaining a corresponding ground truth $\tilde{\rvm}^*$. 
We approximate the max-values $\rvm^*$ empirically using 2000 posterior samples of $\psi$ 
We similarly estimate $\psi$ using 2000 posterior samples. 
We define the optimal action under the posterior model to be the average (with respect to that posterior) optimal action when actions are continuous or $\text{UCB}_0$ when discrete. 
Finally, the regret is computed as the average difference between the true max value (from the true environment and the true optimal action) and the one obtained by applying the estimated optimal action.  
We used 4000 (resp. 2000) true environment realisation for the continuous (resp. discrete) example. 

\subsubsection{Discrete actions example}
\begin{table}[t]
\setlength{\tabcolsep}{6pt}
\renewcommand{\arraystretch}{0.98} 
\center
\caption{Discrete treatments: evaluation metrics of 10D design }
\label{tab:ab_oneseed}
\vspace{5pt}
\begin{tabular}{lccccc}
Method             & EIG estimate      & MSE$(\rvm^*)$      &  Hit rate$(\rvA)$      & Regret     \\
\midrule
$\text{UCB}_{0.0}$ & 1.735 $\pm$ 0.005 & 2.541 $\pm$ 0.104  & 0.513 $\pm$ 0.01 & 1.170 $\pm$ 0.036  \\
$\text{UCB}_{1.0}$ & 2.514 $\pm$ 0.006 & 1.003 $\pm$ 0.043  & 0.496 $\pm$ 0.01 & 1.119 $\pm$ 0.035 \\
$\text{UCB}_{2.0}$ & 2.504 $\pm$ 0.006 & 0.965 $\pm$ 0.045  & 0.497 $\pm$ 0.01 & 1.169 $\pm$ 0.037 \\
Thompson           & 4.607 $\pm$ 0.007 & 0.620 $\pm$ 0.024  & 0.498 $\pm$ 0.01 & 1.112 $\pm$ 0.035 \\
Random             & 3.573 $\pm$ 0.006 & 1.953 $\pm$ 0.070  & 0.503 $\pm$ 0.01 & 1.150 $\pm$ 0.036  \\
\midrule
\textbf{Ours}    & \textbf{4.729} $\pm$\textbf{ 0.009} & \textbf{0.594} $\pm$ \textbf{0.025} & 0.501 $\pm$ 0.01 & 1.152 $\pm$  0.035 \\
\bottomrule
\end{tabular}
\end{table}

\paragraph{Model}
We first give details about the toy model we consider in Figure~\ref{fig:exp_discrete_ab}. Each of the four treatments $\rva = 1, 2, 3, 4$ is a random function with two parameters $\psi_k = (\psi_{k,1}, \psi_{k,2})$ with the following Gaussian priors (parameterised by mean and covariance matrix): 
\begin{align}
    \psi_1 &~ \sim \mathcal{N} \left(\begin{pmatrix} 5.00 \\ 15.0  \end{pmatrix},  
    \begin{pmatrix}
     9.00 & 0 \\ 
     0 & 9.00
    \end{pmatrix} \right) &
    \psi_2 &~ \sim \mathcal{N} \left(\begin{pmatrix} 5.00 \\ 15.0  \end{pmatrix},  
    \begin{pmatrix}
     2.25 & 0 \\ 
     0 & 2.25
    \end{pmatrix} \right) \\
    \psi_3 &~ \sim \mathcal{N} \left(\begin{pmatrix} -2.0 \\ -1.0  \end{pmatrix},  
    \begin{pmatrix}
     1.21 & 0 \\ 
     0 & 1.21
    \end{pmatrix} \right) &
    \psi_4 &~ \sim \mathcal{N} \left(\begin{pmatrix} -7.0 \\ 3.0  \end{pmatrix},  
    \begin{pmatrix}
     1.21 & 0 \\ 
     0 & 1.21
    \end{pmatrix} \right)
\end{align}
and reward (outcome) likelihoods:
\begin{align}
    y | \rvc, \rva, \psi &\sim  \mathcal{N} \left( f(\rvc, \rva, \psi), 0.1 \right) \\
    f(\rvc, \rva, \psi) &= -\rvc^2 + \beta(\rva, \psi)\rvc + \gamma(\rva, \psi) \\
    \gamma &= (\psi_{\rva, 1} + \psi_{\rva, 2} + 18) / 2 \\
    \beta &= (\psi_{\rva, 2} - \gamma + 9) / 3
\end{align}
Intuition about the parameterisation: The first component of each $\psi_i$ defines the mean reward at context $\rvc=-3$, while the second one defines the mean reward at context $\rvc=3$. The reward is then the quadratic equation that passes through those points and the leading coefficient is equal to $-1$. 

\paragraph{Experimental and evaluation contexts} We use experimental and evaluation contexts of the same sizes. The experimental context, $\rvc$ is an equally spaced grid of size 10 between $-3$ and $-1$. We set the evaluation context $\rvc^* = -\rvc$. Figure~\ref{fig:exp_discrete_ab} in the main text visually illustrates this: the $x$-axis of the points in each plot are the experimental contests, while the dashed gray lines are the evaluation contexts.

\paragraph{Further results}
Table \ref{tab:ab_oneseed} shows all the evaluation metrics for the discrete treatment example from the main text. Our method achieves substantially higher EIG and lower MSE for estimating the max-rewards.  On all other metrics, all methods perform similarly. This is to be expected since Treatments 1 and 2 have exactly the same means and due to the way the model was parameterised (by the value of the quadratic at contexts 3 and -3), the probability of the optimal treatment being 1 or 2 is exactly $50\%$ (the hit rate all baselines achieve).
Note that $\text{UCB}_1$ and $\text{UCB}_2$ achieve statistically identical results, which is expected given they select the same designs.

\paragraph{Training stability}  We perform our method with the same hyperparameters but different training seeds and report the mean and standard error in Table~\ref{tab:ab_stability}.

\begin{table}[t]
\setlength{\tabcolsep}{6pt}
\renewcommand{\arraystretch}{0.98} 
\center
\caption{Discrete treatments example: 10D design, stability across training seeds}
\label{tab:ab_stability}
\vspace{5pt}
\begin{tabular}{lccccc}
Method             & EIG estimate      & MSE$(\rvm^*)$          & Hit rate$(\rvA)$  & Regret            \\
\midrule
$\text{UCB}_{0.0}$ & 1.740 $\pm$ 0.003 & 2.709 $\pm$ 0.058  & 0.500 $\pm$ 0.005   & 1.150 $\pm$ 0.017  \\
$\text{UCB}_{1.0}$ & 2.508 $\pm$ 0.002 & 0.993 $\pm$ 0.016  & 0.498 $\pm$ 0.004 & 1.140 $\pm$ 0.007  \\
$\text{UCB}_{2.0}$ & 2.505 $\pm$ 0.006 & 0.991 $\pm$ 0.023   & 0.497 $\pm$ 0.003 & 1.145 $\pm$ 0.015 \\
Thompson           & 4.536 $\pm$ 0.173 & 0.773 $\pm$ 0.127 & 0.500 $\pm$ 0.003 & 1.148 $\pm$ 0.018 \\
Random             & 3.573 $\pm$ 0.333 & 2.369 $\pm$ 0.382   & 0.502 $\pm$ 0.003 & 1.166 $\pm$ 0.008 \\
\midrule
Ours               & \textbf{4.769} $\pm$ \textbf{0.048} & \textbf{0.628} $\pm$ \textbf{0.025}   & 0.502 $\pm$ 0.005 & 1.160 $\pm$ 0.021 \\
\bottomrule
\end{tabular}
\end{table}

\subsubsection{Continuous treatment example}

\begin{table}[t]
\setlength{\tabcolsep}{3pt}
\renewcommand{\arraystretch}{1} 
\centering
\caption{Continuous actions: 40D design training stability. Mean and standard error are reported across 6 different training seeds.}
\label{tab:cts_40d_stability}
\vspace{5pt}
\begin{tabular}{lccccc}
Method             & EIG estimate      & MSE$(\rvm^*)$          & MSE$(\rvA)$       & Regret            \\
\midrule
Random$_{0.2}$     & 5.548 $\pm$ 0.044 & 0.0037 $\pm$ 0.0002    & 0.451 $\pm$ 0.033 & 0.083 $\pm$ 0.004 \\
Random$_{1.0}$     & 5.654 $\pm$ 0.128 & 0.0031 $\pm$ 0.0004    & 0.343 $\pm$ 0.044 & 0.069 $\pm$ 0.006 \\
Random$_{2.0}$     & 5.118 $\pm$ 0.163 & 0.0045 $\pm$ 0.0003    & 0.498 $\pm$ 0.032 & 0.086 $\pm$ 0.004 \\
$\text{UCB}_{0.0}$ & 5.768 $\pm$ 0.002 & 0.0066 $\pm$ 0.0002    & 0.729 $\pm$ 0.022 & 0.082 $\pm$ 0.001 \\
$\text{UCB}_{1.0}$ & 5.892 $\pm$ 0.006 & 0.0031 $\pm$ 0.0001    & 0.354 $\pm$ 0.013 & 0.068 $\pm$ 0.001 \\
$\text{UCB}_{2.0}$ & 5.797 $\pm$ 0.004 & 0.0030 $\pm$ 0.0001    & 0.343 $\pm$ 0.011 & 0.071 $\pm$ 0.001 \\
Thompson           & 6.184 $\pm$ 0.004 & 0.0017 $\pm$ 0.0001    & 0.161 $\pm$ 0.007 & 0.051 $\pm$ 0.001 \\
\midrule
\textbf{Ours}               & \textbf{6.538} $\pm$ \textbf{0.008} & \textbf{0.0013} $\pm$ \textbf{0.0001}  & \textbf{0.131} $\pm$ \textbf{0.006} & \textbf{0.042} $\pm$ \textbf{0.0001}    \\
\bottomrule
\end{tabular}
\end{table}

\paragraph{Model} For the continuous treatment example we use the following model:
\begin{align}
 \text{Prior: }  \quad  &\psi = (\psi_0, \psi_1, \psi_2, \psi_3 ), \quad  \psi_i \sim \text{Uniform}[0.1, 1.1] \; \text{iid} \\
 \text{Likelihood: } \quad &y  |\rvc, \rva, \psi \sim \mathcal{N} (f(\psi, \rva, \rvc), \sigma^2),
\end{align}
where
\begin{equation}
    f(\psi, \rva, \rvc) = \exp\left( -\frac{\big(a - g(\psi, \rvc)\big)^2}{h(\psi, \rvc)} - \lambda a^2 \right) \quad
    g(\psi, \rvc) = \psi_0 + \psi_1 \rvc + \psi_2 \rvc^2 \quad
    h(\psi, c) = \psi_3.
\end{equation}
The parameter $\lambda$ is a cost weight, we set $\lambda = 0.1$ in our experiments.

\paragraph{Experimental and evaluation contexts} The experimental context, $\rvc$ is an equally spaced grid of size $D=40$ (or $20$ or $60$ in Further Results below) between $-3.5$ and and $3.5$. The evaluation context $\rvc^*$ is of size $D^* = D - 1$ and consists of the midpoints of the experimental context. 

\paragraph{Baselines} Since we have a continuous treatment, for the random baseline we consider sampling designs at random from $\mathcal{N}(0, 0.2)$, $\mathcal{N}(0, 1)$ or $\mathcal{N}(0, 2)$, which we denote by $\text{Random}_0.2$, $\text{Random}_1$ and $\text{Random}_2$, respectively.

\paragraph{Training stability} We perform our method with the same hyperparameters but different training seeds and report the mean and standard error in Table~\ref{tab:cts_40d_stability}.

\paragraph{Further results} We report the results of the same experiment, but  with a smaller and larger batch sizes of experimental and evaluation contexts. Table~\ref{tab:cts_20d} shows results for an experimental batch size of 20 contexts to learn about 19 evaluation contexts. 
Finally, Table~\ref{tab:cts_60d} shows results for an experimental batch size of 60 contexts to learn about 59 evaluation contexts.

\begin{table}[t]
\setlength{\tabcolsep}{3pt}
\renewcommand{\arraystretch}{1} 
\centering
\caption{Continuous actions: 20D design to learn about 19 evaluation contexts.}
\label{tab:cts_20d}
\vspace{5pt}
\begin{tabular}{lccccc}
Method             & EIG estimate      & MSE$(\rvm^*)$         & MSE$(\rvA)$       & Regret    \\
\midrule
Random$_{0.2}$     & 4.262 $\pm$ 0.004 & 0.0086 $\pm$ 0.0003   & 1.046 $\pm$ 0.041  & 0.120 $\pm$ 0.002 \\
Random$_{1.0}$     & 4.264 $\pm$ 0.004 & 0.0068 $\pm$ 0.0003   & 0.799 $\pm$ 0.033& 0.114 $\pm$ 0.002 \\
Random$_{2.0}$     & 4.116 $\pm$ 0.003 & 0.0083 $\pm$ 0.0003   & 1.002 $\pm$ 0.044& 0.127 $\pm$ 0.003 \\
$\text{UCB}_{0.0}$ & 5.093 $\pm$ 0.004 & 0.0074 $\pm$ 0.0004   & 0.800 $\pm$ 0.047& 0.097 $\pm$ 0.002   \\
$\text{UCB}_{1.0}$ & 5.040 $\pm$ 0.004 & 0.0072 $\pm$ 0.0004   & 0.764 $\pm$ 0.041& 0.097 $\pm$ 0.002 \\
$\text{UCB}_{2.0}$ & 5.038 $\pm$ 0.004 & 0.0048 $\pm$ 0.0003   & 0.573 $\pm$ 0.033& 0.086 $\pm$ 0.002 \\
Thompson           & 4.924 $\pm$ 0.045 & 0.0055 $\pm$ 0.0004   & 0.547 $\pm$ 0.054 & 0.093 $\pm$ 0.003 \\
\midrule
\textbf{Ours}          & \textbf{5.642} $\pm$ \textbf{0.003} & \textbf{0.0034 } $\pm$ \textbf{0.0002}  & \textbf{0.065} $\pm$ \textbf{0.002} & \textbf{0.034} $\pm$ \textbf{0.027 } \\
\bottomrule
\end{tabular}
\end{table}

\begin{table}[t]
\setlength{\tabcolsep}{3pt}
\renewcommand{\arraystretch}{0.9} 
\centering
\caption{Continuous treatment example: 60D design to learn about 59 evaluation contexts.}
\label{tab:cts_60d}
\vspace{5pt}
\begin{tabular}{lccccc}
Method             & EIG estimate      & MSE$(\rvm^*)$            & MSE$(\rvA)$      & Regret    \\
\midrule
Random$_{0.2}$     & 6.033 $\pm$ 0.003 & 0.0026 $\pm$ 0.0001   & 0.307 $\pm$ 0.019 & 0.068 $\pm$ 0.002 \\
Random$_{1.0}$     & 5.877 $\pm$ 0.004 & 0.0025 $\pm$ 0.0002   & 0.310 $\pm$ 0.023  & 0.064 $\pm$ 0.002 \\
Random$_{2.0}$     & 6.153 $\pm$ 0.003 & 0.0022 $\pm$ 0.0002   & 0.226 $\pm$ 0.019 & 0.055 $\pm$ 0.002 \\
$\text{UCB}_{0.0}$ & 6.106 $\pm$ 0.003 & 0.0056 $\pm$ 0.0004   & 0.586 $\pm$ 0.045 & 0.077 $\pm$ 0.002 \\
$\text{UCB}_{1.0}$ & 6.200 $\pm$ 0.003 & 0.0027 $\pm$ 0.0003   & 0.305 $\pm$ 0.028 & 0.063 $\pm$ 0.002 \\
$\text{UCB}_{2.0}$ & 6.234 $\pm$ 0.003 & 0.0024 $\pm$ 0.0002   & 0.252 $\pm$ 0.024 & 0.064 $\pm$ 0.002 \\
Thompson           & 6.656 $\pm$ 0.018 & 0.0012 $\pm$ 0.0001   & 0.105 $\pm$ 0.008 & 0.043 $\pm$ 0.001 \\
\midrule
\textbf{Ours}     & \textbf{6.932} $\pm$ \textbf{0.003} & \textbf{0.0007} $\pm$ \textbf{0.0001} &  \textbf{0.069} $\pm$ \textbf{0.009} &\textbf{ 0.033} $\pm$ \textbf{0.001} \\
\bottomrule
\end{tabular}
\end{table}

\subsection{Gaussian Processes}
\label{sec:appendix_gp}
We take $\rvc \in [-1, 1]^2$ and $a \in [-1, 1]$. We consider a GP $\psi \sim \mathcal{GP}(0, k)$ where $k$ is a radial basis kernel with length-scale $\tfrac{1}{3}$. Observations are sampled as $y | \rvc, a, \psi \sim N(\psi(\rvc, a), \sigma^2)$.

\begin{figure}
    \centering
	\begin{tikzpicture}
	\node (c1) [draw, circle, fill=lightgray] at (0, 0) {$\rvc$};
    \node (a1) [draw, circle, fill=lightgray] at (2, 0) {$a$};
    \node (y) [draw, circle, fill=lightgray] at (1, -1) {$y$};
	\draw[-{Latex[length=2.5mm]}] (c1) -> (a1);
	\draw[-{Latex[length=2.5mm]}] (c1) -> (y);
	\draw[-{Latex[length=2.5mm]}] (a1) -> (y);
	\end{tikzpicture}
 \caption{The form of the causal graph that generates observational data for \Cref{sec:gp}.}
 \label{fig:gp_graph_tikz}
\end{figure}
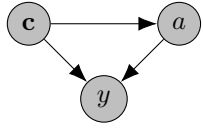

Formally, the confounding bias in observational data arises from the causal graph in \Cref{fig:gp_graph_tikz}.
Concretely, we created 100 initial observational data points; this data was then held fixed across all experiment runs and seeds.
The data was created by sampling $c_i \iid \text{Unif}(-1, 1)$. To create a confounded dataset with $\rvc$ acting as a confounder, we take $a = \text{sign}(c_1) \text{Unif}(0.8, 1)$. 
Finally, we let $y = 1 + \sin(\pi(c_1 - c_2)) - (a - \sin(\pi(c_1 + c_2)))^2$. Note, this function is \emph{not} used for evaluation, instead we sample possible ground truth functions from the GP conditioned on all 100 observational data points. This allows us to validate the robustness of our method to different ground truth functions, and is in keeping with our other experiments.
For the purely random existing data, we resample using exactly the same procedure, except $a \sim \text{Unif}(-1, 1)$ in this case.

The experimental context $\rvC$ was an evenly spaced $7 \times 7$ grid with corners at $(\pm1, \pm1)$. 
The evaluation context $\rvC^*$ was an evenly spaced $4 \times 4$ grid with corners at $(\pm0.8, \pm 0.8)$.

Since it is not possible to sample the infinite dimensional $\psi$, we instead take joint samples of $\psi$ evaluated at $(\rvc_i,a_i)_{i=1}^D, (\rvc^*_j, g_k)_{j=1,k=1}^{D^*,G}$ where $g_1,\dots,g_G$ is a uniform grid covering $[-1, 1]$. From this set of joint samples, we compute each $(m^*_j)_{j=1}^{D^*}$ by maximising over the grid.
Sampling of the multivariate Gaussian admits pathwise derivatives, which we utilise to optimise the design.

For the random baseline, we sample $a \sim \text{Unif}(-1, 1)$. For LP, we follow \citet{groves2018parallelizing} and use a penalization function of the form $1 - k^0(a, a')$. The kernel $k^0$ is chosen to be a RBF kernel with the same length-scale as the kernel of the GP model itself.
We apply the same penalization scheme for the UCB + LP baseline.
Table~\ref{tab:gp_settings} details all the settings used in this experiment.

\begin{figure*}[t]
  \centering
   \begin{subfigure}{0.49\textwidth}
    \includegraphics[width=\hsize]{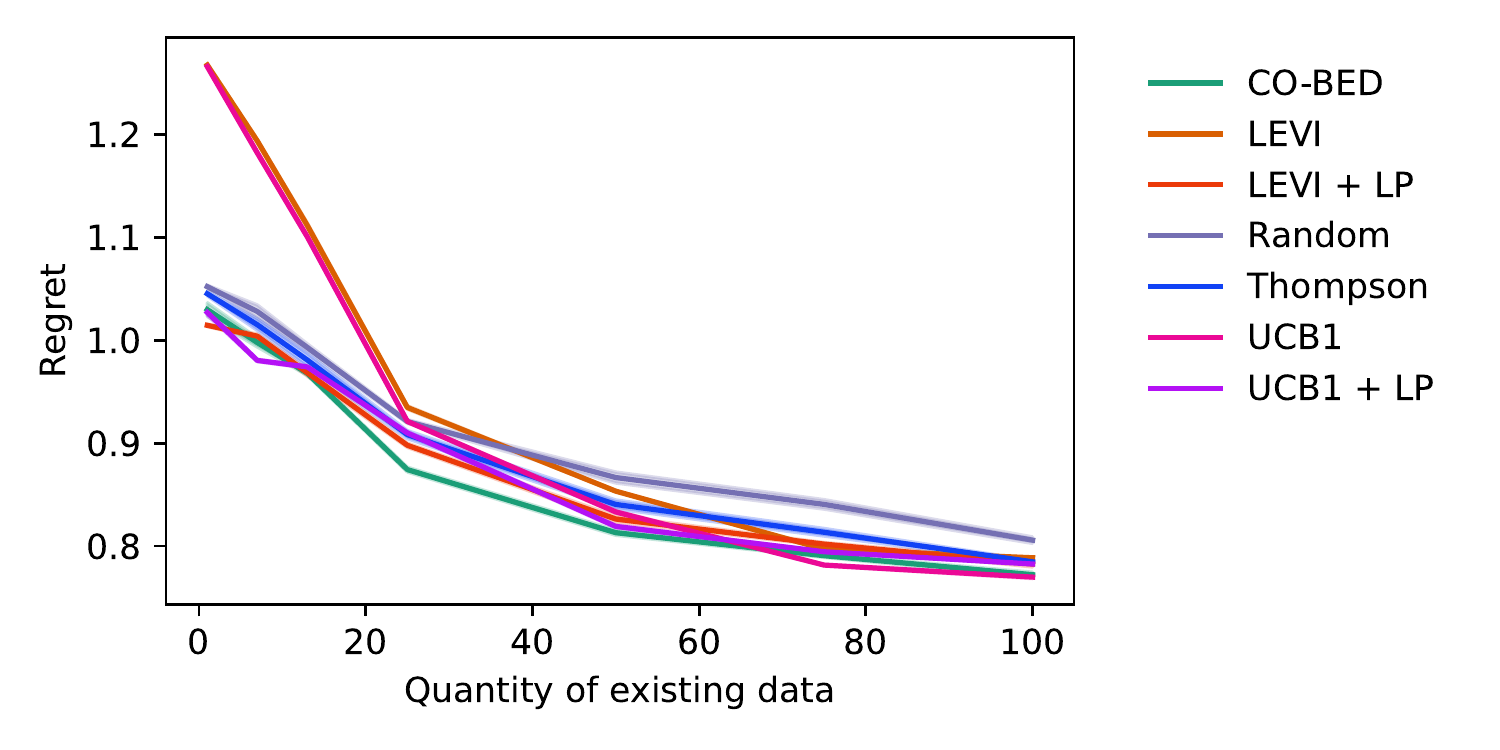}
    \caption{Regret}
   \end{subfigure}
    \hfill
   \begin{subfigure}{0.49\textwidth}
       \includegraphics[width=\hsize]{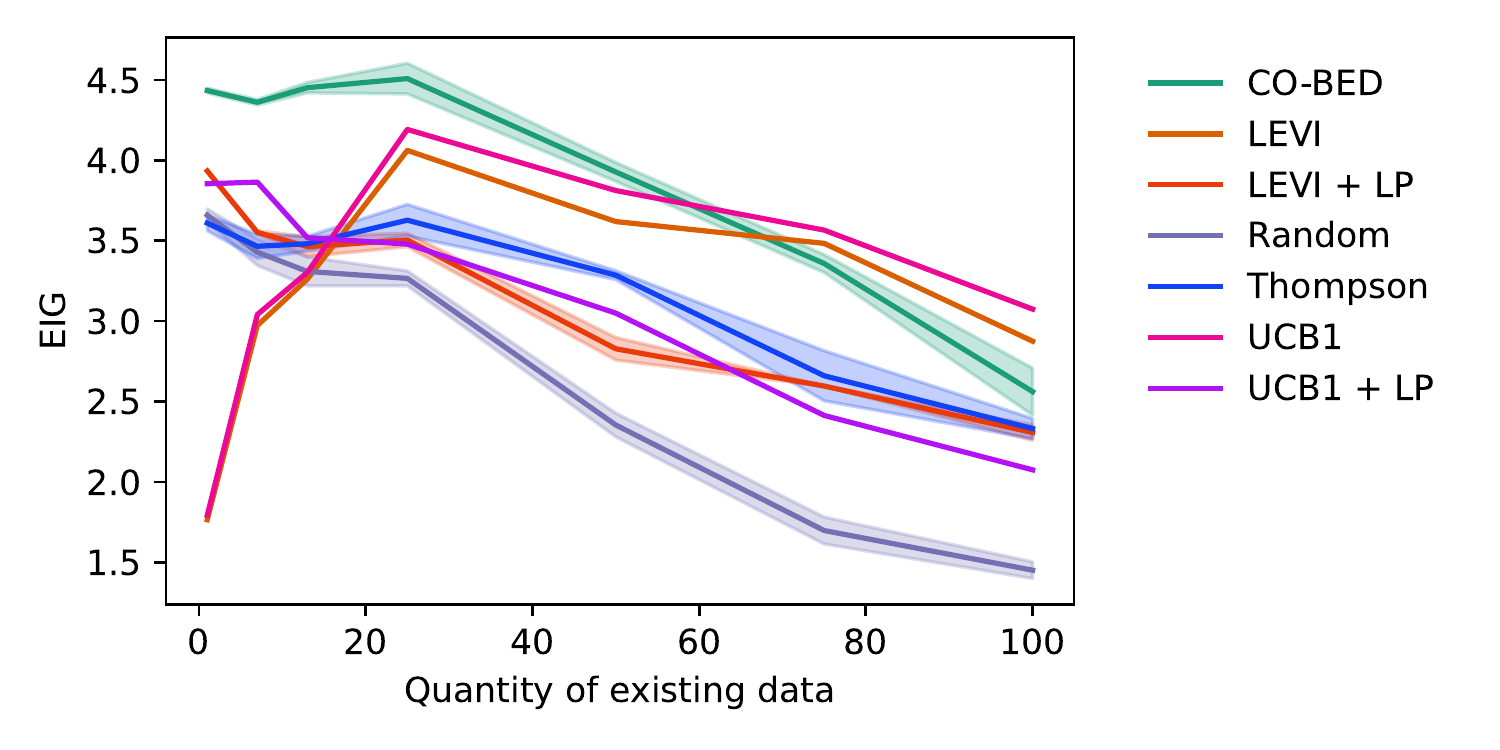}
       \caption{EIG lower bound}
   \end{subfigure}
   \caption{Results from the Gaussian Process example with purely random existing data. We show the mean $\pm 1$ s.e.~from 5 random seeds.}
  \label{fig:gp_rand}
\end{figure*}

We also performed the same experiment, but with observational $\rvc, a$ sampled independently and uniformly (no confounding bias). In this case, the benefits of CO-BED are reduced (\Cref{fig:gp_rand}), although it still does on par with the best of the baselines. Likely, this is because simply `spreading out' designs is a good approach in this case.

\begin{table}[t]
    \centering
    \caption{Parameter settings for the Gaussian Process experiment.}
        \label{tab:gp_settings}
        \vspace{5pt}
    \begin{tabular}{lr}
       Parameter & Value \\
       \hline
       Intrinsic context dimension & 2 \\
       Intrinsic treatment dimension & 1 \\
       Experimental batch size, $D$ & 49 \\
       Evaluation batch size, $D^*$ & 16 \\
       RBF kernel length-scale & $\tfrac{1}{3}$ \\
       Observation noise $\sigma$ & 0.1 \\
       Treatment grid size & 128 \\
       Number of training steps & 50000 \\ 
       Initial learning rate & 0.001 \\
       Learning rate decay factor & 0.96 \\
       Training batch size & 2048 \\
       Critic encoding dimension & 32 \\
       Critic hidden dimension & 256 \\
       Number of ground truth evaluation functions & 3000 \\
    \end{tabular}
\end{table}

\subsection{Contextual Linear Bandits} \label{sec:appendix_exp_bandit}
The random features $\phi(a, c)$ are sampled from $\mathcal{N}(0, \Sigma_{a, c})$, where the covariance matrices are all diagonal 
and of the form $\Sigma_{a, c} = \text{diag}(10^{-9}, \dots, 10^{-9}, 1, 10^{-9}, \dots, 10^{-9} )$, with the position of $1$ specified as follows:
\begin{itemize}
    \item Case  $c = 1$: $\left(\Sigma_{1,1}\right)_{11}$ =  1, $\left(\Sigma_{2,1}\right)_{22}$ =  1, $\left(\Sigma_{3,1}\right)_{33}$ =  1,
    \item Case  $c = 2$: $\left(\Sigma_{1, 2}\right)_{44}$ =  1, $\left(\Sigma_{2, 2}\right)_{11}$ =  1, $\left(\Sigma_{3, 2}\right)_{55}$ =  1,
    \item Case  $c = 3$:  $\left(\Sigma_{1, 3}\right)_{66}$ =  1, $\left(\Sigma_{2, 3}\right)_{77}$ =  1, $\left(\Sigma_{3, 3}\right)_{11}$ =  1,
\end{itemize}

We define the following prior on the parameters $\psi$: $\psi_{1:19} \sim \mathcal{N}(0, 1.0)$ iid, and $\psi_{20} \sim \mathcal{N}(0, 0.1)$.

\paragraph{Experimental and evaluation contexts} The experimental contexts $\rvC$ are sampled uniformly from $\{1, 2 ,3\}$, whilst $\rvC^*= (1, 2 ,3)$.
We varied the design dimension $2, 5, 10, 15, \dots, 60$.

\paragraph{Training details}
All experiments baselines ran for 100K gradient steps, using a batch size of 1024. We used the Adam optimiser \citep{kingma2014adam} with initial learning rate $1e^{-3}$ and exponential learning rate annealing with coefficient 0.96 applied every 1000 steps. We used a separable critic architecture \citep{poole2018variational} with simple MLP encoders with ReLU activations hidden units determined by the size of the design. Concretely, 
we use MLPs with sizes [\texttt{input\_dim}, $2\times$\texttt{input\_dim}, $4\times$\texttt{input\_dim}, \texttt{input\_dim}], where \texttt{input\_dim} is equal to  $D$ (resp. $D^*$) for encoding the outcomes $\rvy$ (resp. max-values $\rvm^*$). 

We added \emph{batch norm} to the critic architecture, which helped to stabilize the optimisation.
Additionally, for the Gumbel--Softmax policy, we started with a temperature $\tau=5.0$ and \texttt{hard=False} constraint.
We applied temperature annealing every 20K steps with a factor 0.5. We switch to \texttt{hard=True} in the last 20K steps of training.

Note: In order to evaluate the EIG of various baselines, we train a critic network for each one of them with the same hyperparameters as above.

\subsection{Unknown Causal Graph}
\label{sec:appendix_causal}

Structural equation modelling \citep{pearl2000models,pearl2009causal} is a mainstay of causal reasoning in statistics.
The causal assumptions of this experiment are captured explicitly in \Cref{fig:causal_graph_tikz}.
This causal graph captures the intuition that $\rvc$ represents contexts that cannot be changed in the short run and $\rva$ represents actions that are directly manipulated. 

We consider a binary context vector $\rvc \in \{0, 1\}^{k}$ indicating which business areas a customer is active in, and treatments $\rva \in [0, 1]^{\ell}$ representing investment in different promotional activities.
The \emph{unobserved} variable $\rvr \in \mathbb{R}^k$ indicates the revenue generated in each business area. The unknown component of the causal graph relates to which treatments effect which revenue streams, with $r_i = c_i\sum_{j=1}^{\ell} G_{ij} \theta_{ij}  a_j$ where $G_{ij}$ is a binary causal graph and $\theta_{ij}$ are linear coefficients. The latent variables of the model are therefore $\psi = \{ G, \theta \}$, each a matrix of shape $k \times \ell$.
In the prior, we sample each component of $G$ independently from $\text{Bernoulli}(3/2k)$ and each component of $\theta \iid \text{HalfNormal}(1)$.
Note that, given our set-up, any sample of $G$ results in the overall causal graph being acyclic, side-stepping some of the complexities of learning distributions over causal graphs more generally \citep{annadani2021variational, geffner2022deep}.
The total cost of treatments is simply $s = \sum_{j=1}^{\ell} a_j$, and the total profit is $y = \sum_{i=1}^k r_i - s + \epsilon$ where $\epsilon$ is sampled $\epsilon \sim N(0.25^2)$. 

At design time, we create a random experimental context $\rvC \in \{0,1\}^{D \times k}$. We fix $D=200, k=8$ and sample each component of the context from $\text{Bernoulli}(0.5)$. This context is sampled once and fixed across seeds and baselines, to focus differences on quality of experimental design.

For CO-BED, we represent $\rva$ in logit space, and use an initialization of $N(0,1)$.
We can compute conditional maximum rewards $m^* \mid \rvc^*, G, \theta$ using the formula
\begin{equation}
    u_j := -1 + \sum_{i=1}^k c_i G_{ij}\theta_{ij}, \qquad a^*_j  = \begin{cases}
    1 \text{ if } u_j > 0, \\
    0 \text{ otherwise}
    \end{cases},
     \qquad m^* = \sum_{j=1}^{\ell} \left(\sum_{i=1}^{k} c_i G_{ij}\theta_{ij}a^*_j \right) - a^*_j.
\end{equation}
For the random baselines, we restricted ourselves to designs placed at the extrema, i.e.~$\rva\in\{0,1\}^{\ell}$. Since the functional relationships in the model are linear, using only extreme values is likely to substantially improve the quality of the baseline. We sampled the baseline design components independently from $\text{Bernoulli}(p)$ for various values of $p$. The UCB designs were computed by first calculating upper confidence bounds on each entry of $G \odot \theta$ (here, $\odot$ indicates the Hadamard, or element-wise, product).

At evaluation time, we created a systematic evaluation context $\rvC \in \mathbb{R}^{D^* \times k}$ that consists of \emph{all} $D^* = 2^k - 1$ non-zero binary context vectors of length $k$. This means that our evaluation is `comprehensive' in the sense that it covers all possible contexts that might be observed. With our choice $k=8$ we have $D^*=255$, which is slightly larger than the total number of experiments $D=200$.

Our standard approach for evaluation is to first sample a ground truth $\psi^*$ from the prior, sample experimental outcomes $\rvc \sim p(\rvy|\rvC,\rvA,\psi^*)$, and then compute the posterior $p(\psi|\rvy,\rvC,\rvA)$.
However, in this case, computing this posterior constitutes solving a partial causal discovery problem on an adjacency matrix of size $k \times \ell$; doing this accurately is an area of ongoing active research. Instead, we substitute the posterior calculation for a point estimate. We begin by observing that $y + s = \sum_{i,j} c_iG_{ij}\theta_{ij}a_j + \epsilon$ can be interpreted as a linear model with coefficient equal to the pointwise product $G \odot \theta$ and covariates given by the outer product of $\rva$ and $\rvc$. 
We therefore estimate $G \odot \theta$ by fitting a Lasso \citep{tibshirani1996regression} to the data that was sampled under $\psi^*$. The Lasso has a long history in causal discovery, and is considered a robust approach to estimating the causal parents \citep{friedman2008sparse,shortreed2017outcome}. We select the Lasso penalty weight $\alpha$ using cross validation independently for each run.
Our results appear to accord very well with the ground truth graphs and optimal actions, particularly at lower dimensions, indicating that our approach to approximate causal discovery is suitable in this case.

\begin{figure*}[t]
  \centering
  \begin{subfigure}{0.49\textwidth}
    \includegraphics[width=\hsize]{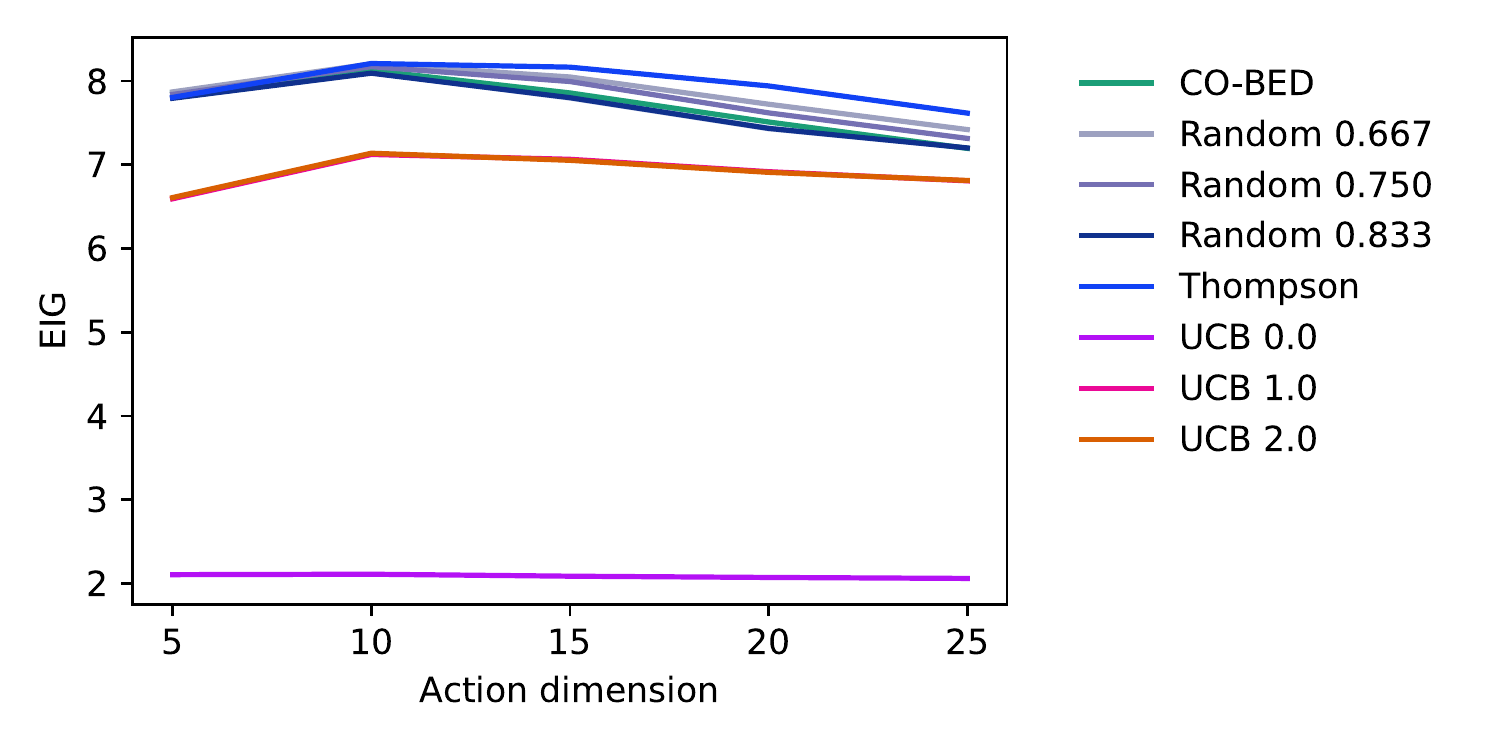}
    \caption{CMV-EIG}
   \end{subfigure}
    \hfill
   \begin{subfigure}{0.49\textwidth}
    \includegraphics[width=\hsize]{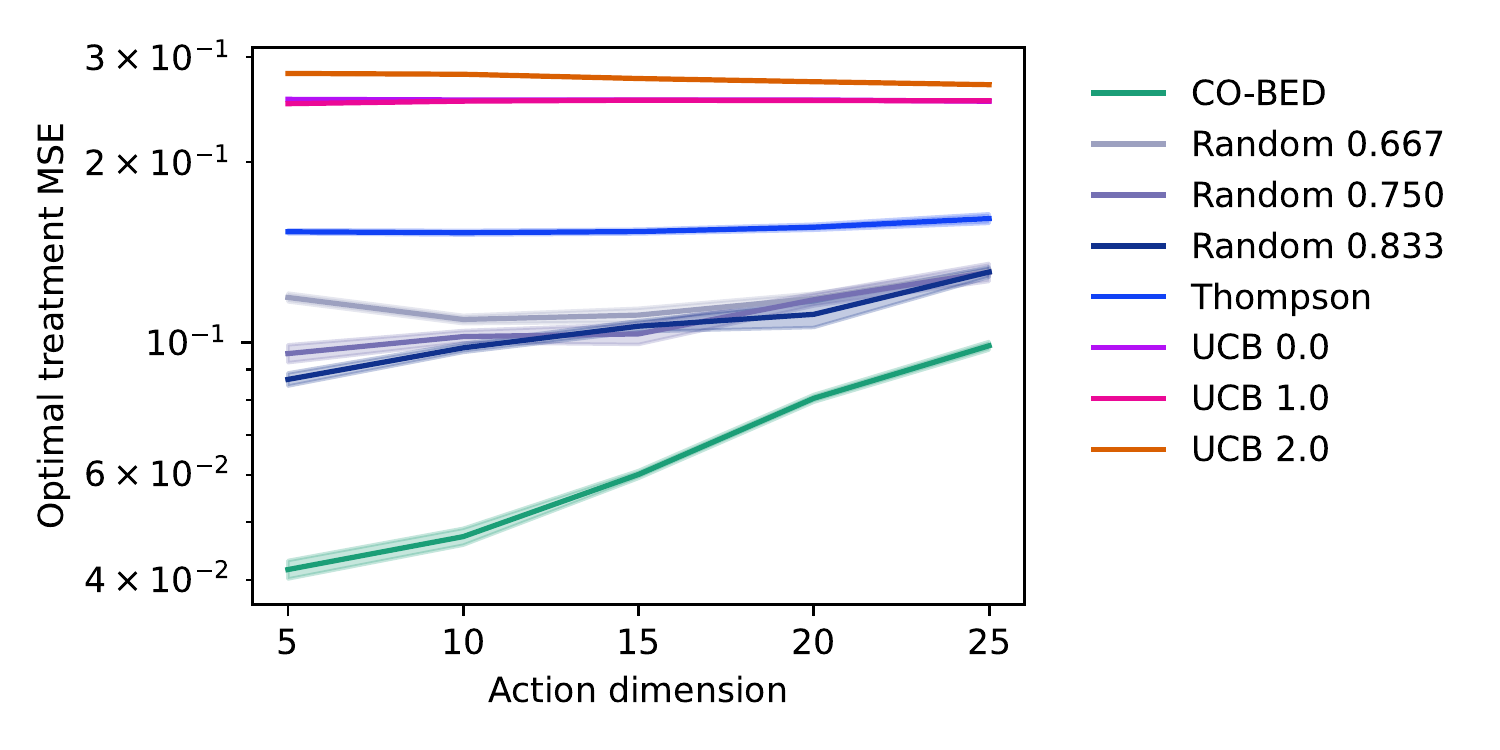}
    \caption{Optimal treatment MSE}
   \end{subfigure}
    \hfill
   \begin{subfigure}{0.49\textwidth}
    \includegraphics[width=\hsize]{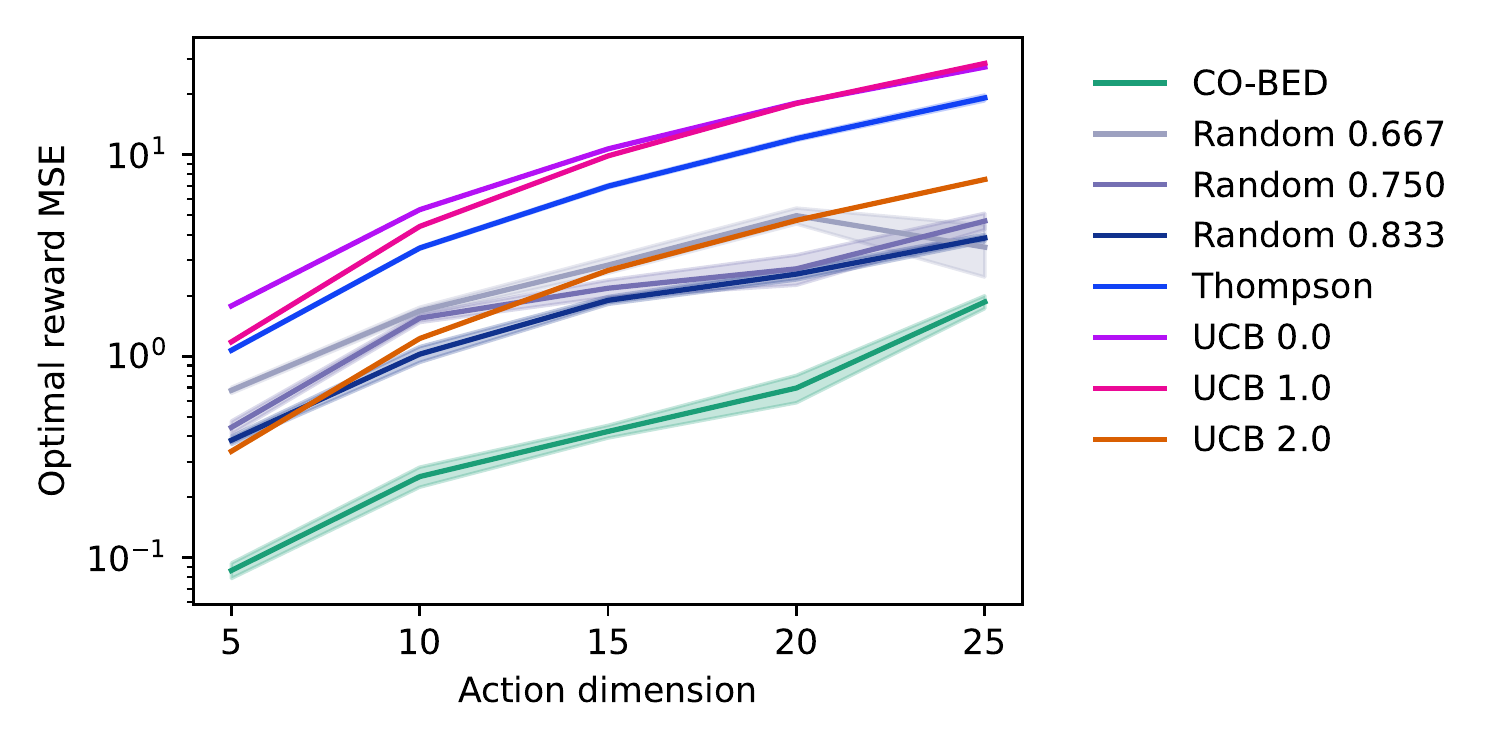}
    \caption{Optimal reward MSE}
   \end{subfigure}
   \caption{Additional metrics from the Unknown Causal Graph experiment. Plots show the mean $\pm 1$ s.e.~from 5 seeds.}
  \label{fig:causal_graph_extra}
\end{figure*}

\Cref{fig:causal_graph_extra} shows additional metrics from our experiment. Interestingly, we see that CO-BED does \emph{not} outperform other methods on the EIG metric, despite this being the objective that is directly optimised. 
We believe that this finding is related to the EIG objective for these large scale experiments being near to its maximum value of $\log(4096) = 8.318$. It is not possible to exceed this bound without increasing the batch size further. 
Secondly, for the baselines, the critic can fully adapt to a fixed design throughout 400000 training iterations, whereas for CO-BED, the critic has to adapt to a design that changes during training, and is therefore likely to improve further with yet longer training.
This experiment shows convincingly that the InfoNCE objective can give good training gradients for experimental designs \emph{even when it is close to saturation at} $\log (\text{batch size})$.
Finally, \Cref{tab:causalgraph_settings} details the settings used in our experiment.

\begin{table}[t]
\caption{Parameter settings for the Unknown Causal Graph experiment.}
\label{tab:causalgraph_settings}
\vspace{5pt}
    \centering
    \begin{tabular}{lr}
       Parameter & Value \\
       \hline
       Intrinsic context dimension, $k$ & 8 \\
       Intrinsic treatment dimension, $\ell$ & $5, 10, 15, 20, 25$ \\
       Experimental batch size, $D$ & 200 \\
       Experimental context sampling probability & 0.5 \\
       Evaluation batch size, $D^*$ & 255 \\
       Observation noise scale & 0.25 \\
       Graph prior probability & $\tfrac{3}{16}$ \\
       Linear coefficient prior & $\text{HalfNormal}(1)$ \\
       Random baseline probability $p$ & $0.667, 0.75, 0.833$ \\
       Number of training steps & 400000 \\ 
       Initial learning rate (critic) & $3 \times 10^{-6}$ \\
       Initial learning rate (design) & $3 \times 10^{-5}$ \\
       Learning rate decay factor & 0.998 \\
       Training batch size & 4096 \\
       Critic encoding dimension & 256 \\
       Critic hidden dimension & 512 \\
       Number of ground truth evaluation functions & 10000 \\
    \end{tabular}
    
\end{table}

\twocolumn

\end{document}